\documentclass{article} 
\usepackage{iclr2027_conference,times}

\iclrfinalcopy

\usepackage{amsmath,amsfonts,bm}

\def\eqref#1{equation~\ref{#1}}

\def\1{\bm{1}}

\DeclareMathAlphabet{\mathsfit}{\encodingdefault}{\sfdefault}{m}{sl}
\SetMathAlphabet{\mathsfit}{bold}{\encodingdefault}{\sfdefault}{bx}{n}

\usepackage[T1]{fontenc}
\usepackage{hyperref}
\usepackage{url}
\usepackage{graphicx}
\usepackage{float}
\usepackage{placeins}
\usepackage{algorithm}
\usepackage{algpseudocode}
\usepackage{xcolor}
\usepackage{bm}
\usepackage{booktabs}
\usepackage{multirow}
\usepackage{makecell}
\usepackage{array}
\usepackage{caption}

\usepackage{amsmath}
\usepackage{amssymb}
\usepackage{tabularx}
\usepackage{fvextra}
\usepackage[most]{tcolorbox}
\tcbuselibrary{listings,breakable,skins}
\newsavebox{\iterativebaselinebox}

\usepackage{titletoc}
\titlecontents{appsection}
    [16pt]
    {\addvspace{8pt}\bfseries}
    {\contentslabel{16pt}}
    {}
    {\hfill\contentspage}

\titlecontents{appsubsection}
    [38pt]
    {\addvspace{2pt}\normalfont}
    {\contentslabel{23pt}}
    {}
    {\titlerule*[0.7pc]{.}\contentspage}

\titlecontents{appsubsubsection}
    [70pt]
    {\addvspace{1pt}\normalfont}
    {\contentslabel{32pt}}
    {}
    {\titlerule*[0.7pc]{.}\contentspage}

\definecolor{PromptCountdown}{HTML}{738DA2}
\definecolor{PromptCountdownBG}{HTML}{F4F7F9}

\definecolor{PromptGSM}{HTML}{5F8F98}
\definecolor{PromptGSMBG}{HTML}{F2F8F8}

\definecolor{PromptCode}{HTML}{668B78}
\definecolor{PromptCodeBG}{HTML}{F3F8F5}

\definecolor{PromptStory}{HTML}{81759A}
\definecolor{PromptStoryBG}{HTML}{F7F5FA}

\definecolor{PromptChem}{HTML}{A77474}
\definecolor{PromptChemBG}{HTML}{FAF5F5}

\definecolor{PromptMath500}{HTML}{7A7898}
\definecolor{PromptMath500BG}{HTML}{F6F5FA}

\definecolor{PromptOlympiad}{HTML}{667A9A}
\definecolor{PromptOlympiadBG}{HTML}{F3F5F9}

\lstdefinestyle{promptstyle}{
  basicstyle=\ttfamily\scriptsize,
  breaklines=true,
  breakatwhitespace=false,
  columns=fullflexible,
  keepspaces=true,
  showstringspaces=false,
  upquote=true,
  tabsize=2,
  aboveskip=0pt,
  belowskip=0pt,
  literate=
    {→}{{$\rightarrow$}}1
    {’}{{\textquoteright}}1
}

\newtcblisting{promptbox}[3]{
  enhanced,
  breakable,
  listing only,
  listing engine=listings,
  listing options={style=promptstyle},
  title={#1},
  fonttitle=\bfseries,
  coltitle=white,
  colbacktitle=#2,
  colframe=#2,
  colback=#3,
  boxrule=0.55pt,
  arc=1.5mm,
  outer arc=1.5mm,
  left=2.8mm,
  right=2.8mm,
  top=1.8mm,
  bottom=1.8mm,
  before skip=0.8em,
  after skip=0.8em
}

\definecolor{modularblue}{RGB}{225,240,252}

\newcommand{\ModState}[1]{%
  \State
  \colorbox{modularblue}{%
    \parbox{\dimexpr0.88\linewidth-2\fboxsep\relax}{#1}%
  }%
}

\newcommand{\ModStateTop}[1]{%
  \State
  \colorbox{modularblue}{%
    \parbox{\dimexpr0.88\linewidth-2\fboxsep\relax}{#1}%
  }%
  \vspace{-0.22em}%
}

\title{Modular Norm RandOpt: Population-Efficient Ensembling through Architecture-Aware Perturbations}

\author{Kirato Yoshihara \\
The University of Osaka \\
\texttt{kiratoyoshihara@gmail.com} \\
\And
Hiroaki Hamade \\
The University of Osaka \\
\texttt{hiroakihamade@gmail.com} \\
}

\begin{document}

\maketitle
\lhead{Modular Norm RandOpt}
\raggedbottom

\begin{abstract}
RandOpt samples weight-perturbed language models and ensembles
top-ranked candidates through plurality voting, but its global
perturbation scale ignores heterogeneous module geometry.
We propose \mbox{\textbf{\emph{Modular Norm RandOpt}}}, an
architecture-aware sampling method using module-wise natural norms
and calibrated scales while preserving selection and voting.
It outperforms RandOpt using $3\times$ fewer candidates on Countdown
and at least $12\times$ fewer on GSM8K, with corresponding
wall-clock savings.
Evaluations across seven tasks and three Qwen scales
($0.5$B--$3$B) show higher mean accuracy than RandOpt on Countdown,
GSM8K, and MATH-500 at every scale.
The gains extend to Llama 3.2 $3$B and Gemma 3 $4$B on Countdown
and GSM8K.
On Qwen2.5-1.5B, our ensembles also achieve higher mean accuracy
than iterative baselines on both tasks at comparable main-run
evaluation budgets.
On GSM8K, a tail-density diagnostic implies only a
$1.2$--$1.8\times$ candidate reduction, while most ensemble
improvement is associated with more favorable correct-expert support.
These results highlight perturbation geometry as a key design
choice for population-efficient, gradient-free search around
pretrained models.
\end{abstract}

\vspace{-0.9em}
\begin{figure}[!ht]
    \centering

    \captionsetup{
        width=1.30\textwidth,
        justification=raggedright,
        singlelinecheck=false
    }

    \makebox[\textwidth][c]{%
        \includegraphics[width=1.30\textwidth]{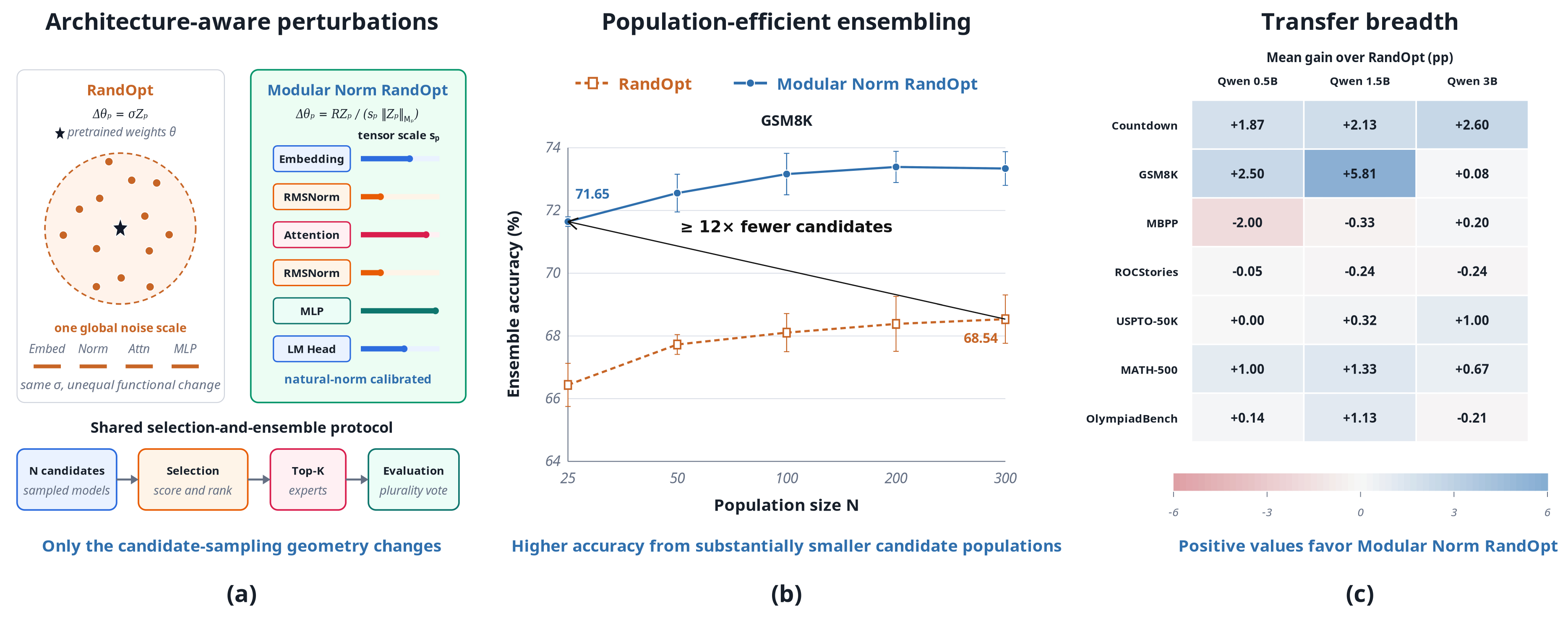}%
    }

    \vspace{-0.6em}

    \caption{
    \textbf{Modular Norm RandOpt enables architecture-aware,
    population-efficient ensembling.}
    \textbf{(a)} Modular Norm RandOpt replaces isotropic perturbations with
    module-wise natural-norm and architecture-aware scaling, while keeping
    selection and ensembling unchanged.
    \textbf{(b)} On GSM8K with $K=25$, it exceeds RandOpt using
    $\geq 12\times$ fewer candidates; error bars show one sample standard
    deviation over three seeds.
    \textbf{(c)} The calibrated geometry transfers across seven tasks and
    three Qwen model scales.
    }

    \label{fig:overview}
    \vspace{-0.5em}
\end{figure}

\par\smallskip
{\small
\hypersetup{hidelinks}
\noindent\makebox[\linewidth][c]{%
  \makebox[0.42\linewidth][r]{%
    \href{https://kiratoyoshihara.github.io/Modular-Norm-RandOpt-page/}
         {\textcolor{blue}{Project page}}%
  }%
  \quad\quad
  \makebox[0.42\linewidth][l]{%
    \href{https://github.com/kiratoyoshihara/Modular-Norm-RandOpt}
         {\textcolor{blue}{Code}}%
  }%
}
\par}

\section{Introduction}
\label{intro}
Modern language models have gained broad capabilities through scaling model size,
data, and computation \citep{GPT-3, PaLM, Chinchilla}, with open model families
demonstrating strong performance across architectures and scales
\citep{Llama, Llama2, Mistral-7B, OLMo, qwen2024qwen25, llama3_2024, gemma2025gemma3}.
After pretraining, models are commonly adapted through supervised, parameter-efficient,
preference-based, or feedback-driven methods
\citep{RLHF, LoRA, chung2024scaling, Self-instruct, Constitutional-AI, DPO}.

As a gradient-free alternative, RandOpt samples $N$ random weight perturbations
     around a pretrained model, evaluates them on a small selection set, and
     ensembles the top $K$ by plurality vote \citep{neural-thickets}.
It can be competitive with more involved post-training procedures,
consistent with evidence that useful task specialists are dense near
well-pretrained weights \citep{neural-thickets}.
However, every candidate must be evaluated before selection, so search cost
grows directly with $N$. We therefore ask:
\emph{can equally strong ensembles be obtained from substantially smaller
candidate populations without changing RandOpt's selection or ensembling?}

We argue that a key opportunity lies in the geometry used to sample candidates.
RandOpt perturbs the full parameter vector with isotropic Gaussian noise controlled
by a global scale, whereas Transformer models are compositions of heterogeneous
modules with distinct parameterizations and functional roles
\citep{transformer, ba2016layer, RMSNorm, shazeer2020glu, RoPE, GQA}.
Equal Euclidean perturbation magnitudes therefore need not induce comparable
functional changes across the network.
The modular norm formalizes an architecture-aware geometry for optimization by
assigning modules natural norms and composing them according to network structure
\citep{modular-norm}.
This motivates using the same geometric principle to shape RandOpt's candidate
distribution.

We introduce \textbf{Modular Norm RandOpt}, which replaces RandOpt's
isotropic sampling with architecture-aware perturbations normalized by
module-wise natural norms and calibrated modular scales.
Candidate evaluation, top-$K$ selection, and plurality voting remain unchanged,
isolating candidate-sampling geometry as the intervention
(Figure~\ref{fig:overview}(a)).

This change substantially improves population efficiency.
On Countdown, Modular Norm RandOpt with $N=100$ matches or exceeds RandOpt
with $N=300$; on GSM8K \citep{GSM8K}, $N=25$ exceeds RandOpt with $N=300$
at both $K=10$ and $K=25$, corresponding to $3\times$ and at least
$12\times$ fewer candidates.
Across seven tasks
\citep{GSM8K, MBPP, ROCStories, USPTO-50K, MATH-500, olympiadbench}
and three Qwen2.5 scales \citep{qwen2024qwen25}, our method achieves higher mean performance than RandOpt in $14$ of
$21$ settings and ties in one. The gains also extend to Llama 3.2 $3$B \citep{llama3_2024} and Gemma 3 $4$B \citep{gemma2025gemma3}, with additional
cross-family evaluation on OLMo 3 $7$B-Instruct \citep{Olmo-3} on Countdown and GSM8K.

On GSM8K, high-reward tail enrichment predicts only a
$1.2$--$1.8\times$ candidate reduction, far below the observed
$\geq12\times$ reduction.
Our ensemble-level analysis shows that most of the GSM8K accuracy gain
arises from a more favorable distribution of correct-expert support
under plurality voting.

\section{Background and Related Work}
\label{sec:background}

\subsection{RandOpt and Gradient-Free Post-Training}

Evolution strategies \citep{salimans2017evolution,ES-Scale},
black-box prompt tuning \citep{black-box-tuning}, and zeroth-order
fine-tuning \citep{MeZO,ZO-Finetuner} perform iterative optimization.
CMA-ES adapts search covariances \citep{hansen2001cmaes}, while
natural evolution strategies update search distributions using
natural gradients \citep{wierstra2014natural}.
RandOpt instead ensembles selected candidates from a fixed population,
with parallelizable candidate evaluations \citep{neural-thickets}.
CoRP consolidates rewarded perturbations into one model
\citep{zhang2026consolidating}; we instead modify candidate generation while
preserving RandOpt's selection and voting.

\subsection{Perturbation Geometry and Modular Structure}

Modular norms normalize updates according to network structure
\citep{modular-norm}, while modular duality constructs duality maps
from layer-wise norms \citep{bernstein2025modular}.
Manifold-constrained GPT-2 experiments also suggest different geometry
preferences for attention and MLP modules \citep{yoshihara2026different}.
Safe Mutations rescales mutations using output sensitivities to weights
\citep{lehman2018safe}; PATS adds sensitivity-dependent noise during
language-model fine-tuning \citep{zhang2022pats}.
\citet{kim2026not} examine perturbation dimension, subspace, and norm
under fixed scoring and voting.
We combine natural norms, architecture-based allocation, and
module-input Jacobian estimates to fix the sampling geometry before
candidate search.

\subsection{Weight-Space Diversity and Ensembling}

Deep and snapshot ensembles combine independently trained predictors
or training checkpoints \citep{Deep-ensembles,Snapshot-ensembles}.
SWAG samples from a Gaussian weight posterior approximation fitted to
SGD iterates \citep{maddox2019swag}, while PEP perturbs trained weights
\citep{PEP}.
RandOpt additionally ranks candidates on a selection set
\citep{neural-thickets}.
Self-consistency samples reasoning paths in output space
\citep{Self-consistency}, whereas RandOpt and our method perturb weights.
Ensemble theory highlights trade-offs between member errors and diversity
\citep{wood2023unified}, motivating our analysis of correct-expert
support and voting beyond individual candidate scores.

\section{Modular Norm RandOpt}
\label{sec:method}

RandOpt samples candidate models by adding isotropic Gaussian
perturbations to pretrained parameters, ranks them on a selection set,
and ensembles the top-$K$ candidates \citep{neural-thickets}.
We retain this selection and ensembling procedure and modify only the
candidate sampling geometry.
\textbf{Modular Norm RandOpt} replaces the global isotropic scale with
parameter specific perturbations determined by natural norms and
sensitivity calibrated recursive modular scales.
Figure~\ref{fig:overview}(a) illustrates this distinction.

\subsection{Architecture-Aware Perturbation Geometry}
\label{sec:architecture_aware_geometry}

Let $\bm{\theta}=\{\theta_p\}_{p\in\mathcal{P}}$ denote the
pretrained parameters, where $\mathcal{P}$ is the set of physical
parameter tensors to be perturbed.
Each tensor $p$ is represented by a leaf module $\mathsf{M}_p$
in the model's module tree, with
\begin{equation}
    m_p:=\mathsf{M}_p.\mathrm{mass},
    \qquad
    \|\cdot\|_{\mathsf{M}_p}:=\mathsf{M}_p.\mathrm{norm}.
    \label{eq:leaf_module_attributes}
\end{equation}
Here $m_p>0$ is a fixed architecture-based weight that allocates
perturbation magnitude across tensors, not the tensor's parameter
count, and $\|\cdot\|_{\mathsf{M}_p}$ is its role-specific norm.
The scale $s_p>0$ combines this mass allocation with a calibrated
sensitivity correction
(Section~\ref{sec:sensitivity_calibrated_scales}).
The resulting max-form geometry on the full perturbation is
\begin{equation}
    \|\Delta\bm{\theta}\|_{\mathsf{M}}
    :=
    \max_{p\in\mathcal{P}}
    s_p\|\Delta\theta_p\|_{\mathsf{M}_p},
    \label{eq:modular_max_geometry}
\end{equation}
where $\mathsf{M}$ denotes the root model module.
For candidate $i$, we generate a standard Gaussian noise tensor
$Z_{i,p}\sim\mathcal{N}(\mathbf{0},\mathbf{I})$ with the same shape as $\theta_p$ and set

\begin{equation}
    \Delta\theta_{i,p}
    =
    R\frac{Z_{i,p}}
    {s_p\|Z_{i,p}\|_{\mathsf{M}_p}},
    \qquad
    \theta'_{i,p}=\theta_p+\Delta\theta_{i,p},
    \label{eq:modular_perturbation}
\end{equation}

Here $R>0$ is the global modular radius.
Each active tensor has natural-norm magnitude $R/s_p$, so larger
$s_p$ implies a smaller perturbation.
Hence
$s_p\|\Delta\theta_{i,p}\|_{\mathsf{M}_p}=R$ and
$\|\Delta\bm{\theta}_i\|_{\mathsf{M}}=R$.

We use role-specific natural norms: maximum-row $\ell_2$ for
embeddings, spectral norm for linear maps, and $\ell_\infty$ for
one-dimensional parameters.
Appendix~\ref{app:module_norms} gives the complete mapping,
including fused projections and fallback cases.

\begin{algorithm}[t]
\caption{\textbf{Modular Norm RandOpt.}
Highlighted lines are the architecture-aware normalization and scaling steps
that differ from RandOpt.}
\label{alg:modular_randopt}
\begin{algorithmic}[1]

\Require pretrained parameters $\bm{\theta}$,
selection set $\mathcal{D}_{\mathrm{sel}}$,
population size $N$,
ensemble size $K$,
radius $R$,
fixed scales $\{s_p\}_{p\in\mathcal{P}}$

\For{$i=1,\ldots,N$}

    \State For each $p\in\mathcal{P}$, sample
    $Z_{i,p}\sim\mathcal{N}(\mathbf{0},\mathbf{I})$

    \ModStateTop{%
    Compute $\|Z_{i,p}\|_{\mathsf{M}_p}$ for every
    $p\in\mathcal{P}$%
    }

    \ModState{%
    Set
    $\displaystyle
    \Delta\theta_{i,p}
    \gets
    R\frac{Z_{i,p}}
    {s_p\|Z_{i,p}\|_{\mathsf{M}_p}}$
    for all $p\in\mathcal{P}$%
    }

    \State Construct candidate parameters
    $\theta'_{i,p}\gets\theta_p+\Delta\theta_{i,p}$
    for all $p\in\mathcal{P}$

    \State
    $\mathrm{score}_i
    \gets
    \mathrm{Evaluate}(\bm{\theta}'_i,\mathcal{D}_{\mathrm{sel}})$

\EndFor

\State
$\mathcal{I}_{\mathrm{top}}
\gets
\mathrm{TopK}
\left(
\{\mathrm{score}_i\}_{i=1}^{N},K
\right)$

\For{test input $x$}

    \State
    $\mathcal{A}(x)
    \gets
    \{\mathrm{Generate}(\bm{\theta}'_i,x):
      i\in\mathcal{I}_{\mathrm{top}}\}$

    \State
    $\widehat{y}(x)
    \gets
    \mathrm{PluralityVote}\bigl(\mathcal{A}(x)\bigr)$

\EndFor

\State \Return $\widehat{y}$

\end{algorithmic}
\end{algorithm}

\subsection{Sensitivity-Calibrated Recursive Modular Scales}
\label{sec:sensitivity_calibrated_scales}

The scale $s_p$ combines the mass allocation induced by the module tree with
a layer-relative functional-sensitivity correction.  For calibration example
$e$, Transformer layer $\ell$, and functional-map index $\tau$, we estimate the
largest local amplification of a unit Euclidean activation perturbation,

\begin{equation}
    \widehat{\gamma}_{e,\ell,\tau}
    \approx
    \max_{\|v\|_2=1}
    \left\|
        J_{f_{\ell,\tau}}(x_{e,\ell,\tau})v
    \right\|_2,
    \qquad
    J_f(x):=\frac{\partial f(x)}{\partial x}.
    \label{eq:local_sensitivity_main}
\end{equation}
Here $f_{\ell,\tau}$ is the functional map indexed by $\tau$
in layer $\ell$, and $x_{e,\ell,\tau}$ is its input for
calibration example $e$.
Let $\mathcal{I}_\ell$ denote the calibration examples assigned to layer
$\ell$.  We aggregate the estimates in log space,
\begin{equation}
    \Gamma_{\ell,\tau}
    =
    \exp\!\left[
        Q_{0.9}
        \left(
            \left\{
                \log\max
                \bigl(
                    \widehat{\gamma}_{e,\ell,\tau},
                    10^{-12}
                \bigr)
            \right\}_{e\in\mathcal{I}_\ell}
        \right)
    \right].
    \label{eq:aggregated_sensitivity_main}
\end{equation}
Here $Q_{0.9}$ denotes the empirical 90th percentile.
With the root scale set to one and all functional sensitivities in the
recursive modular composition rule replaced by one, the mass ratios telescope
along each root-to-leaf path:
\begin{equation}
    s_p^{\mathrm{base}}
    =
    \frac{m_{\mathsf{M}}}{m_p},
    \qquad
    m_{\mathsf{M}}
    :=
    \mathsf{M}.\mathrm{mass}
    =
    \sum_{p\in\mathcal{P}}m_p.
    \label{eq:base_modular_scale}
\end{equation}
Parameters with $m_p=0$ are excluded from perturbation. The aggregated sensitivities and projection gains define a role-specific raw
correction $r_p^{\mathrm{raw}}$. 
For an active tensor $p$ within a Transformer layer, let $\ell(p)$
denote its layer and $\mathcal{P}_\ell$ the set of active tensors
in layer $\ell$. Its layer-relative correction and final scale are
\begin{equation}
    \rho_p
    =
    \operatorname{clip}
    \left(
        \frac{r_p^{\mathrm{raw}}}
        {\operatorname{median}
         \{r_q^{\mathrm{raw}}:q\in\mathcal{P}_{\ell(p)}\}},
        \frac{1}{2},
        2
    \right),
    \qquad
    s_p=s_p^{\mathrm{base}}\rho_p.
    \label{eq:layer_relative_correction}
\end{equation}

Combining the scale construction with
Equation~\ref{eq:modular_perturbation}, the perturbation magnitude is
\begin{equation}
    \|\Delta\theta_{i,p}\|_{\mathsf{M}_p}
    =
    \frac{R}{s_p}
    =
    R\frac{m_p}{m_{\mathsf{M}}}\frac{1}{\rho_p}.
    \label{eq:perturbation_allocation}
\end{equation}
Thus, the mass fraction sets the architecture-based baseline
perturbation magnitude, while $\rho_p$ provides a layer-relative
sensitivity adjustment bounded to a factor of two in either direction. For active tensors outside the repeated Transformer
layers, we set $\rho_p=1$.
All scales are fixed before candidate search and shared across
candidates. Appendix~\ref{app:sensitivity_calibration} gives the calibration
protocol, mass allocation, and construction of $r_p^{\mathrm{raw}}$.

\begin{table}[H]
\centering
\caption{\textbf{Transfer and runtime results.}
MN RandOpt denotes Modular Norm RandOpt;
RMSNorm-only perturbs only RMSNorm weights.
Values are mean $\pm$ sample SD over seeds 42--44.
\textbf{(a)} Qwen2.5 task/scale transfer at $N=100$, $K=25$.
The Modular Norm RandOpt sensitivity profile and radius are determined
on Countdown and transferred unchanged across tasks.
\textbf{(b)} Model-family transfer at $N=100$, $K=25$, with radius
selected on Countdown and fixed on GSM8K.
Scores in (a,b) are accuracy (\%), except USPTO-50K
(balanced accuracy).
\textbf{(c)} Wall-clock minutes at $K=25$, comparing RandOpt
$N=300$ with Modular Norm RandOpt $N=100$ on Countdown and $N=25$ on GSM8K.
Bold marks the highest mean transfer score within each model/task
(all tied methods) or lower runtime.}
\label{tab:transfer_runtime}
\vspace{0.25em}

\begin{minipage}[t]{\linewidth}
\centering
\textbf{(a) Qwen task and scale transfer}
\par\vspace{0.25em}
\scriptsize
\setlength{\tabcolsep}{1.9pt}
\setlength{\medmuskip}{2mu}
\renewcommand{\arraystretch}{1.12}

\begin{tabular*}{\linewidth}{@{\extracolsep{\fill}}ll*{7}{r}@{}}
\toprule
Model & Method & Countdown & GSM8K & MBPP & ROCStories
& USPTO-50K & MATH-500 & \shortstack{Olympiad\\Bench} \\
\midrule

\multirow{3}{*}{0.5B}
& RandOpt
& $7.73\pm0.83$
& $53.35\pm0.58$
& $\boldsymbol{26.73}\pm0.99$
& $0.82\pm0.03$
& $\boldsymbol{10.00}\pm0.00$
& $38.67\pm0.58$
& $14.42\pm1.36$ \\
& RMSNorm-only
& $1.27\pm0.31$
& $45.36\pm0.24$
& $26.27\pm0.81$
& $\boldsymbol{0.89}\pm0.01$
& $\boldsymbol{10.00}\pm0.00$
& $29.67\pm0.33$
& $11.32\pm0.32$ \\
& MN RandOpt
& $\boldsymbol{9.60}\pm2.80$
& $\boldsymbol{55.85}\pm0.76$
& $24.73\pm2.37$
& $0.77\pm0.05$
& $\boldsymbol{10.00}\pm0.00$
& $\boldsymbol{39.67}\pm1.76$
& $\boldsymbol{14.56}\pm1.27$ \\

\addlinespace[0.25em]
\multirow{3}{*}{1.5B}
& RandOpt
& $35.67\pm1.22$
& $68.39\pm1.12$
& $\boldsymbol{49.47}\pm0.81$
& $\boldsymbol{6.89}\pm0.12$
& $11.34\pm0.60$
& $58.33\pm1.20$
& $28.62\pm0.74$ \\
& RMSNorm-only
& $24.00\pm0.35$
& $62.04\pm0.39$
& $48.40\pm0.35$
& $6.65\pm0.01$
& $10.77\pm0.04$
& $54.33\pm0.67$
& $23.84\pm0.37$ \\
& MN RandOpt
& $\boldsymbol{37.80}\pm0.87$
& $\boldsymbol{74.20}\pm0.32$
& $49.13\pm0.81$
& $6.65\pm0.08$
& $\boldsymbol{11.65}\pm0.57$
& $\boldsymbol{59.67}\pm0.33$
& $\boldsymbol{29.75}\pm0.76$ \\

\addlinespace[0.25em]
\multirow{3}{*}{3B}
& RandOpt
& $49.53\pm0.64$
& $86.08\pm0.23$
& $60.13\pm0.46$
& $\boldsymbol{14.22}\pm0.05$
& $12.95\pm0.61$
& $68.33\pm0.58$
& $\boldsymbol{41.91}\pm0.80$ \\
& RMSNorm-only
& $41.60\pm0.60$
& $82.97\pm0.12$
& $57.93\pm0.31$
& $13.90\pm0.05$
& $\boldsymbol{15.11}\pm0.40$
& $63.11\pm0.69$
& $36.43\pm0.64$ \\
& MN RandOpt
& $\boldsymbol{52.13}\pm0.99$
& $\boldsymbol{86.15}\pm0.16$
& $\boldsymbol{60.33}\pm0.61$
& $13.98\pm0.06$
& $13.95\pm0.47$
& $\boldsymbol{69.00}\pm1.20$
& $41.70\pm0.32$ \\

\bottomrule
\end{tabular*}
\end{minipage}

\par\vspace{0.75em}

\begin{minipage}[t]{0.49\linewidth}
\vspace{0pt}
\centering
\textbf{\strut (b) Model-family transfer}
\par\vspace{0.25em}
\scriptsize
\setlength{\tabcolsep}{2pt}
\renewcommand{\arraystretch}{1.04}

\begin{tabular*}{\linewidth}{@{\extracolsep{\fill}}llrr@{}}
\toprule
Model & Method & Countdown & GSM8K \\
\midrule

\multirow{2}{*}{\shortstack[l]{Llama 3.2\\3B}}
& RandOpt
& $40.93\pm1.03$
& $80.95\pm0.23$ \\
& MN RandOpt
& $\boldsymbol{44.60\pm1.74}$
& $\boldsymbol{81.45\pm0.31}$ \\

\addlinespace[0.25em]
\multirow{2}{*}{\shortstack[l]{Gemma 3\\4B}}
& RandOpt
& $73.27\pm0.83$
& $87.11\pm0.20$ \\
& MN RandOpt
& $\boldsymbol{73.47\pm1.29}$
& $\boldsymbol{87.79\pm0.35}$ \\

\addlinespace[0.25em]
\multirow{2}{*}{\shortstack[l]{OLMo 3\\7B}}
& RandOpt
& $86.07\pm1.03$
& $\boldsymbol{89.21\pm0.09}$ \\
& MN RandOpt
& $\boldsymbol{86.40\pm0.40}$
& $88.73\pm0.24$ \\

\bottomrule
\end{tabular*}
\end{minipage}%
\hfill%
\begin{minipage}[t]{0.49\linewidth}
\vspace{0pt}
\centering
\textbf{\strut (c) Wall-clock time}
\par\vspace{0.25em}
\scriptsize
\setlength{\tabcolsep}{2pt}

\renewcommand{\arraystretch}{1.50}

\begin{tabular*}{\linewidth}{@{\extracolsep{\fill}}llrrr@{}}
\toprule
Task & Method & $N$ & Search & Total \\
\midrule

\multirow{2}{*}{Countdown}
& RandOpt & 300
& $45.86\pm0.40$
& $52.15\pm0.52$ \\
& MN RandOpt & 100
& $\boldsymbol{14.82\pm0.22}$
& $\boldsymbol{21.07\pm0.35}$ \\

\addlinespace[0.25em]
\multirow{2}{*}{GSM8K}
& RandOpt & 300
& $29.47\pm0.24$
& $37.79\pm0.30$ \\
& MN RandOpt & 25
& $\boldsymbol{2.43\pm0.16}$
& $\boldsymbol{10.85\pm0.17}$ \\

\bottomrule
\end{tabular*}
\end{minipage}

\end{table}

\section{Experimental Setup}
\label{sec:experiments}

\paragraph{Models and tasks.}
We evaluate primarily on Qwen2.5-Instruct models at $0.5$B, $1.5$B,
and $3$B parameters \citep{qwen2024qwen25}, and test model-family
transfer on Llama 3.2 $3$B \citep{llama3_2024}, Gemma 3 $4$B
\citep{gemma2025gemma3}, and OLMo 3 $7$B-Instruct \citep{Olmo-3}. Our seven tasks span arithmetic and mathematical reasoning, code generation, commonsense reasoning, and reaction prediction: Countdown, GSM8K \citep{GSM8K}, MBPP \citep{MBPP}, ROCStories \citep{ROCStories}, USPTO-50K \citep{USPTO-50K}, MATH-500 \citep{MATH-500}, and OlympiadBench \citep{olympiadbench}. We report accuracy except for USPTO-50K, for which we use balanced accuracy.

\paragraph{Evaluation protocol.}
Unless otherwise stated, results are averaged over seeds $42$, $43$, and $44$, and we report the mean and sample standard deviation. Transfer experiments use population size $N=100$ and ensemble size $K=25$. For population scaling, we evaluate prefixes
\[
N\in\{25,50,100,200,300\}
\]
of the same $N_{\max}=300$ candidate population within each run and report $K\in\{10,25\}$. Candidate ranking uses a fixed selection set of $200$ examples, while final performance is measured on the corresponding held-out evaluation set.

\paragraph{Calibration and transfer.}
For task transfer, we calibrate a profile on Countdown separately for each
Qwen model scale and reuse it unchanged on target tasks without recalibration.

\paragraph{Implementation.}
RandOpt, Modular Norm RandOpt, and iterative ES use
\texttt{bfloat16}, whereas MeZO and ZO-Finetuner use
\texttt{float16} following training-only numerical-stability
checks (Appendix~\ref{app:zo_baselines}).
All methods use greedy decoding and the same task-specific
completion caps. Task-specific prompting, scoring rules,
preprocessing, and additional implementation details are
provided in Appendix~\ref{app:implementation}.
For the wall-clock comparison, we use Qwen2.5-1.5B with $K=25$ on
Countdown and GSM8K under identical hardware and inference settings,
and report both candidate-search and end-to-end execution time over three seeds.

\section{Population-Efficient Ensembling and Transfer}
\label{sec:main_results}

\subsection{Population Scaling} \label{sec:population_scaling} 

\begin{figure}[!ht] 
\centering 
\includegraphics[width=1.0\linewidth]{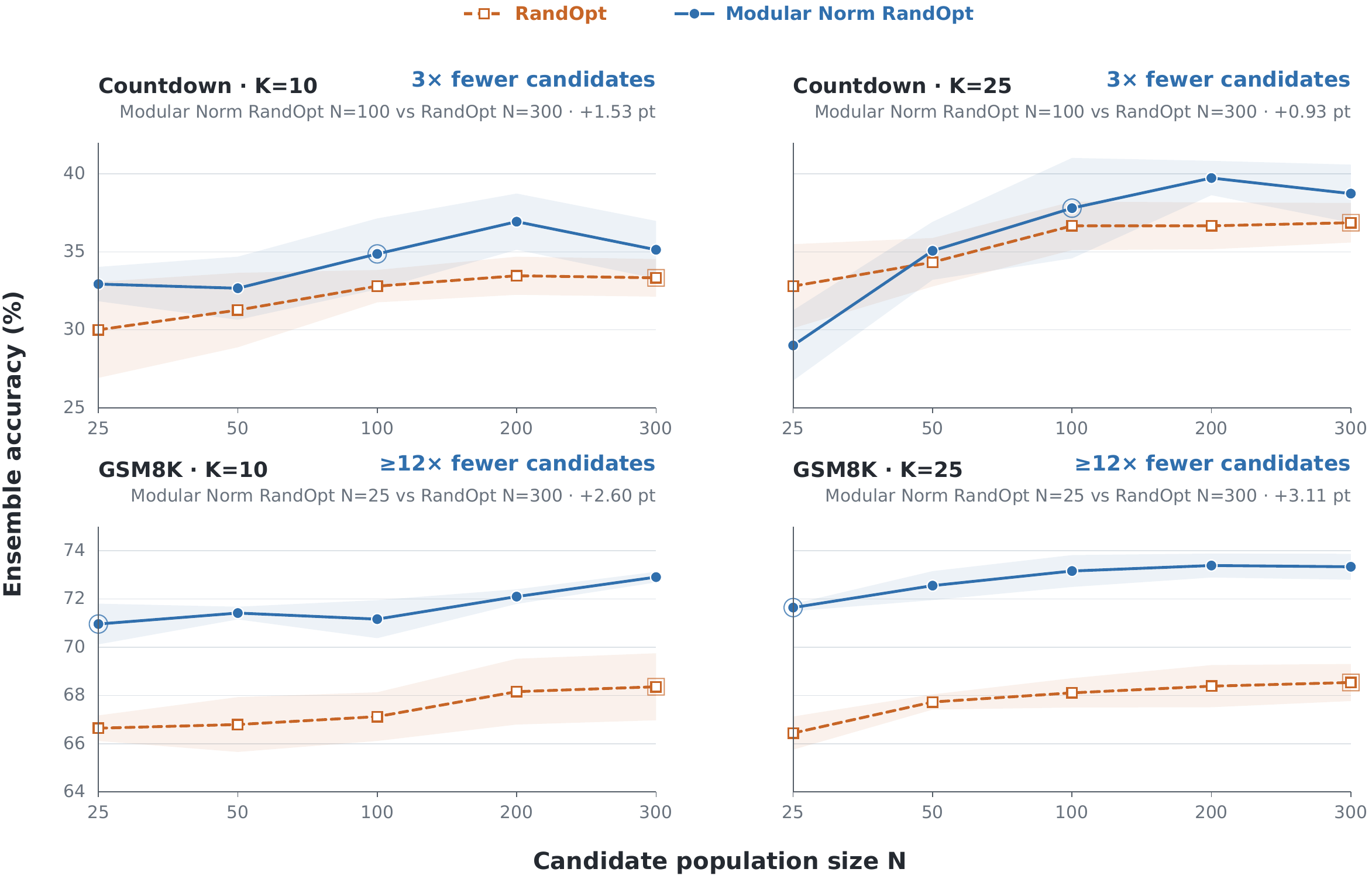} \caption{ 
    \textbf{Population scaling on Countdown and GSM8K.} At both $K=10$ and $K=25$, Modular Norm RandOpt exceeds the $N=300$ RandOpt reference using $N=100$ candidates on Countdown and $N=25$ on GSM8K. Curves show means and shaded regions show one sample standard deviation over seeds $42$, $43$, and $44$. Population sizes are nested prefixes of the same $N_{\max}=300$ population within each run. } 
    \label{fig:population_scaling} 
\end{figure}

\noindent
\begin{minipage}[t]{0.55\linewidth}
\vspace{0pt}

We first compare ensemble accuracy as the candidate budget increases. At both $K=10$ and $K=25$, Modular Norm RandOpt exceeds the
$N=300$ RandOpt reference using $N=100$ on Countdown and $N=25$
on GSM8K (Figure~\ref{fig:population_scaling}), corresponding to
$3\times$ and at least $12\times$ fewer candidates.

\par\smallskip
In a separate comparison at $N=3{,}000$, $K=25$ for both methods,
Modular Norm RandOpt also achieves higher mean accuracy on both tasks
(Table~\ref{tab:large_population}).

\end{minipage}%
\hfill
\begin{minipage}[t]{0.42\linewidth}
\vspace{0pt}
\centering
\footnotesize

\captionsetup{
    font=footnotesize,
    skip=2pt,
    justification=raggedright,
    singlelinecheck=false,
    hypcap=false
}
\captionof{table}{%
\textbf{Large-population comparison.}
Qwen2.5-1.5B-Instruct with $N=3{,}000$, $K=25$ for both methods.
Accuracy (\%): mean $\pm$ sample SD over seeds 42--44.
Bold marks the higher mean.
}
\label{tab:large_population}

\setlength{\tabcolsep}{2pt}
\setlength{\medmuskip}{2mu}
\renewcommand{\arraystretch}{1.08}

\begin{tabular*}{\linewidth}{@{\extracolsep{\fill}}lrr@{}}
\toprule
Method & Countdown & GSM8K \\
\midrule
RandOpt
& $35.29\pm1.60$
& $69.95\pm1.11$ \\
MN RandOpt
& $\boldsymbol{38.44}\pm1.20$
& $\boldsymbol{74.45}\pm0.95$ \\
\bottomrule
\end{tabular*}

\end{minipage}
\par

\FloatBarrier

\subsection{Transfer Across Tasks and Model Scales}
\label{sec:task_transfer}

We next test transfer beyond the calibration task.
For each Qwen2.5 scale, we calibrate the profile on Countdown
and reuse it unchanged across target tasks.
At $N=100$ and $K=25$, Modular Norm RandOpt achieves higher
mean scores than RandOpt in $14$ of $21$ settings and ties
in one (Table~\ref{tab:transfer_runtime}(a)).
The full method also matches or exceeds the RMSNorm-only
baseline's reported mean in $18$ of $21$ settings, suggesting
that RMSNorm weight perturbations alone do not explain the gains.
Improvements over RandOpt span all three scales on Countdown,
GSM8K, and MATH-500, but benefits remain task dependent,
with failures on ROCStories and the smaller MBPP models.

\subsection{Transfer Across Model Families}
\label{sec:model_family_transfer}

To test transfer beyond Qwen, we evaluate Llama 3.2 $3$B,
Gemma 3 $4$B, and OLMo 3 $7$B-Instruct
(Table~\ref{tab:transfer_runtime}(b)).
Modular Norm RandOpt improves over RandOpt in all four Llama
and Gemma comparisons, with OLMo extending evaluation to $7$B.

\begin{table}[t]
\centering

\captionsetup{
    font=footnotesize,
    skip=3pt,
    justification=raggedright,
    singlelinecheck=false
}
\caption{\textbf{Runtime efficiency, iterative baselines, and ablations.}
Accuracy (\%) and speedups are mean $\pm$ sample SD over seeds 42--44.
\textbf{(a)} Modular Norm RandOpt versus iterative baselines.
$N$: cumulative perturbation evaluations;
Eval.\ red.: main-run budget ratio
(ES/method; Appendix~\ref{app:es-budget}).
\textbf{(b)} Modular Norm RandOpt speedups over RandOpt for
Table~\ref{tab:transfer_runtime}(c).
\textbf{(c)} Countdown ablations on Qwen2.5-1.5B-Instruct
at $N=100$, $K=25$.
Bold marks higher mean accuracy or fewer evaluations than ES in (a),
speedups in (b), and the highest mean accuracy in (c).}
\label{tab:efficiency_controls}

\label{tab:runtime_speedup}
\label{tab:iterative-es}
\label{tab:ablations}

\small


\sbox{\iterativebaselinebox}{%
\begin{minipage}[t]{0.53\linewidth}
\vspace{0pt}
\centering

\textbf{\strut (a) Comparison with iterative baselines}
\par\vspace{0.15em}

\footnotesize
\setlength{\tabcolsep}{2pt}
\setlength{\medmuskip}{2mu}
\setlength{\aboverulesep}{1pt}
\setlength{\belowrulesep}{1pt}
\renewcommand{\arraystretch}{1.00}

\begin{tabular*}{\linewidth}{@{\extracolsep{\fill}}lrrrr@{}}
\toprule
Method & $N$ & $K$ & Acc.\ (\%) & Eval.\ red. \\
\midrule

\multicolumn{5}{@{}l}{Countdown} \\

Iterative ES & 3,000 & 1
& $35.67\pm5.08$ & $1.00\times$ \\

ZO-Finetuner & 3,000 & 1
& $29.56\pm0.82$ & $0.99\times$ \\

MeZO & 3,000 & 1
& $29.22\pm1.93$ & $0.99\times$ \\

MN RandOpt & 3,000 & 1
& $16.18\pm1.08$ & $1.00\times$ \\

MN RandOpt & 2,820 & 25
& $\boldsymbol{39.09}\pm1.56$ & $1.00\times$ \\

MN RandOpt & 100 & 25
& $\boldsymbol{38.40}\pm2.71$ & $\boldsymbol{10.46\times}$ \\

\midrule

\multicolumn{5}{@{}l}{GSM8K} \\

Iterative ES & 3,000 & 1
& $73.11\pm0.52$ & $1.00\times$ \\

ZO-Finetuner & 3,000 & 1
& $72.53\pm0.58$ & $0.99\times$ \\

MeZO & 3,000 & 1
& $71.72\pm0.27$ & $0.99\times$ \\

MN RandOpt & 3,000 & 1
& $64.42\pm0.29$ & $1.00\times$ \\

MN RandOpt & 2,841 & 25
& $\boldsymbol{74.43}\pm0.81$ & $1.00\times$ \\

MN RandOpt & 100 & 25
& $\boldsymbol{74.20}\pm0.32$ & $\boldsymbol{11.35\times}$ \\

\bottomrule
\end{tabular*}

\end{minipage}%
}


\noindent
\usebox{\iterativebaselinebox}%
\hfill%
\begin{minipage}[t][%
    \dimexpr\ht\iterativebaselinebox+\dp\iterativebaselinebox\relax
][t]{0.44\linewidth}
\vspace{0pt}
\centering


\textbf{\strut (b) Runtime speedups ($\times$)}
\par\vspace{0.15em}

\footnotesize
\setlength{\tabcolsep}{2pt}
\setlength{\medmuskip}{2mu}
\setlength{\aboverulesep}{1pt}
\setlength{\belowrulesep}{1pt}
\renewcommand{\arraystretch}{1.00}

\begin{tabular*}{\linewidth}{@{\extracolsep{\fill}}lrr@{}}
\toprule
Task & Search & End-to-end \\
\midrule

Countdown
& $\boldsymbol{3.10\pm0.03}$
& $\boldsymbol{2.48\pm0.02}$ \\

GSM8K
& $\boldsymbol{12.15\pm0.84}$
& $\boldsymbol{3.49\pm0.08}$ \\

\bottomrule
\end{tabular*}

\vfill


\textbf{\strut (c) Countdown ablations}
\par\vspace{0.15em}

\footnotesize
\setlength{\tabcolsep}{2pt}
\setlength{\medmuskip}{2mu}
\setlength{\aboverulesep}{1pt}
\setlength{\belowrulesep}{1pt}
\renewcommand{\arraystretch}{1.00}

\begin{tabular*}{\linewidth}{@{\extracolsep{\fill}}lr@{}}
\toprule
Variant & Accuracy (\%) \\
\midrule

Full (ours)
& $\boldsymbol{37.80}\pm0.87$ \\

No sensitivity
& $36.87\pm1.62$ \\

RMSNorm-only scale correction
& $36.80\pm1.64$ \\

No recursive scaling
& $35.80\pm1.22$ \\

Frobenius norm
& $34.40\pm0.72$ \\

Attention only
& $24.00\pm1.56$ \\

MLP only
& $23.73\pm0.64$ \\

Pretrained base
& $12.40\pm0.00$ \\

\bottomrule
\end{tabular*}

\end{minipage}

\end{table}

\subsection{Wall-clock efficiency}
\label{sec:wallclock}

Table~\ref{tab:transfer_runtime}(c) compares wall-clock times for
RandOpt at $N=300$ and Modular Norm RandOpt at $N=100$ on Countdown and
$N=25$ on GSM8K, with matched hardware, inference settings,
and $K=25$.
Speedups (Table~\ref{tab:efficiency_controls}(b)) closely track
candidate reductions during search but are smaller end-to-end
due to fixed-ensemble evaluation and other population-independent costs.

\subsection{Comparison with iterative baselines}
\label{sec:iterative-es}

We compare Modular Norm RandOpt with iterative ES
\citep{ES-Scale}, MeZO \citep{MeZO}, and a task-adapted
ZO-Finetuner \citep{ZO-Finetuner} on Countdown and GSM8K
(Table~\ref{tab:efficiency_controls}(a)).
For the ES comparison, we adjust $N$ to match main-run
evaluation budgets rather than candidate counts
(Appendix~\ref{app:es-budget}).
MeZO and ZO-Finetuner each perform 1,500 two-sided updates,
using 600,000 training model--prompt evaluations;
their checkpoint-selection costs are additionally counted.
Settings are provided in Appendix~\ref{app:zo_baselines}.
Our $K=25$ ensembles achieve higher mean accuracy than ES, 
MeZO, and ZO-Finetuner on both tasks at ES-matched budgets. 
At $N=100$, $K=25$, Modular Norm RandOpt uses approximately $10.6\times$ and $11.5\times$ fewer model--prompt evaluations 
than MeZO or ZO-Finetuner on Countdown and GSM8K, respectively (Appendix~\ref{app:es-budget}).
At $N=3{,}000$ and $K=1$, however, our method underperforms
iterative ES on both tasks.
These results support population-efficient ensembling rather
than superior single-model optimization.


\subsection{Ablations and Controls}
\label{sec:ablations}

Table~\ref{tab:efficiency_controls}(c) compares module coverage,
norm choice, and scaling components.
The full method achieves the highest mean accuracy.
Removing sensitivity correction or recursive scaling lowers
the mean by $0.93$ and $2.00$ percentage points, respectively,
while the Frobenius-norm control is $3.40$ points lower.
Attention- and MLP-only controls retain full-method perturbation
scales and mask other perturbations to zero without renormalization,
yet yield much lower accuracy.
RMSNorm-only scale correction remains close to the full method,
with a $1.00$-point mean difference.
Together, these controls favor combining broad module coverage
with calibrated modular scaling.

\section{Mechanistic Analysis: Why Does It Work?}
\label{sec:mechanism}
We next examine how the perturbation geometry changes the candidate
population and why this can improve the selected ensemble. The
analysis separates candidate-level reward effects from changes that
emerge only after selection and voting.

\subsection{Candidate-Level Effects}
\label{sec:candidate_effects}

\begin{figure}[!htbp]
    \centering
    \vspace{-0.4em}

    \includegraphics[width=0.96\linewidth]{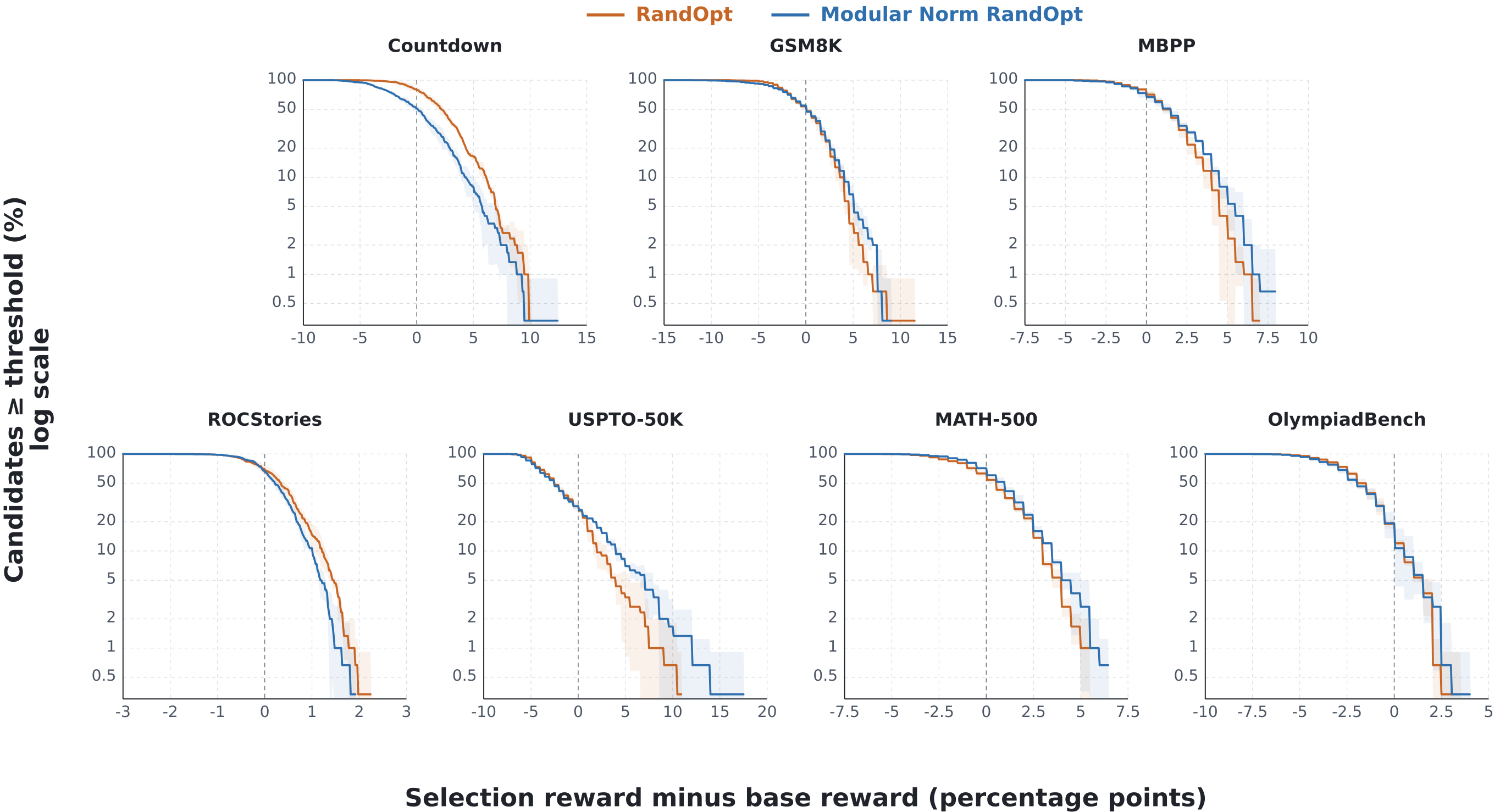}

    \captionsetup{
        font=footnotesize,
        skip=1pt
    }
    \caption{%
    \textbf{Selection-reward distributions across tasks.}
    Curves show the mean percentage of candidates exceeding each
selection-reward-gain threshold; shading shows mean $\pm$ one sample SD
across seeds 42--44, not confidence intervals. Bands are clipped to
the displayed probability range.
    }
    \label{fig:candidate_rewards}

    \vspace{-0.6em}
\end{figure}

Figure~\ref{fig:candidate_rewards} shows that candidate-reward distributions
do not shift uniformly. Modular Norm RandOpt has a more favorable high-reward
tail on GSM8K, whereas RandOpt has a heavier tail on Countdown despite the
stronger Modular Norm RandOpt ensemble. Candidate quality alone therefore
cannot fully explain the gains.

\FloatBarrier

\subsection{Tail Enrichment Does Not Explain the Full Gain}
\label{sec:tail_analysis}

\noindent
\begin{minipage}[t]{0.57\linewidth}
\vspace{0pt}
\small
\raggedright

We quantify this discrepancy on GSM8K, where Modular Norm RandOpt
produces a denser high-reward tail.

\par\smallskip

Figure~\ref{fig:tail_efficiency} uses a tail-based required-population
diagnostic: for each seed, we compute the population needed to obtain
at least $10$ above-threshold candidates with probability at least $0.9$
and average the paired ratios (Appendix~\ref{app:tail_efficiency}).
The resulting $1.2$--$1.8\times$ reduction is far below the observed
$\geq12\times$ reduction, where Modular Norm RandOpt with $N=25$ exceeds
RandOpt with $N=300$ at $K=10$.

\par

\captionsetup{
    font=footnotesize,
    skip=4pt,
    justification=raggedright,
    singlelinecheck=false,
    hypcap=false
}
\captionof{figure}{%
  \textbf{Tail enrichment does not explain the full gain on GSM8K.}
  Both panels use threshold $\tau$ on selection-reward gain over
  the pretrained base(percentage points).
  \textbf{Top:} $p(\tau)$ is the fraction of candidates with gain
  at least $\tau$; vertical bands show mean $\pm$ sample SD of the
  top-$10$ selection cutoffs.
  \textbf{Bottom:} Mean paired-seed ratio
  $N^{\mathrm{req}}_{\mathrm{RandOpt},s}(\tau)/
   N^{\mathrm{req}}_{\mathrm{MN},s}(\tau)$,
  where $N^{\mathrm{req}}$ is the smallest population giving at least
  $10$ above-threshold candidates with probability at least $0.9$
  under a binomial model.
  The dashed line marks the observed $\geq12\times$ candidate reduction.
  Curves show means over seeds 42--44; ribbons show mean $\pm$
  one sample SD, not confidence intervals.
  Both vertical axes are logarithmic; bands are clipped to the plotting limits.
}
\label{fig:tail_efficiency}

\end{minipage}%
\hfill
\begin{minipage}[t]{0.42\linewidth}
\vspace{0pt}
\raggedleft

\includegraphics[width=\linewidth]{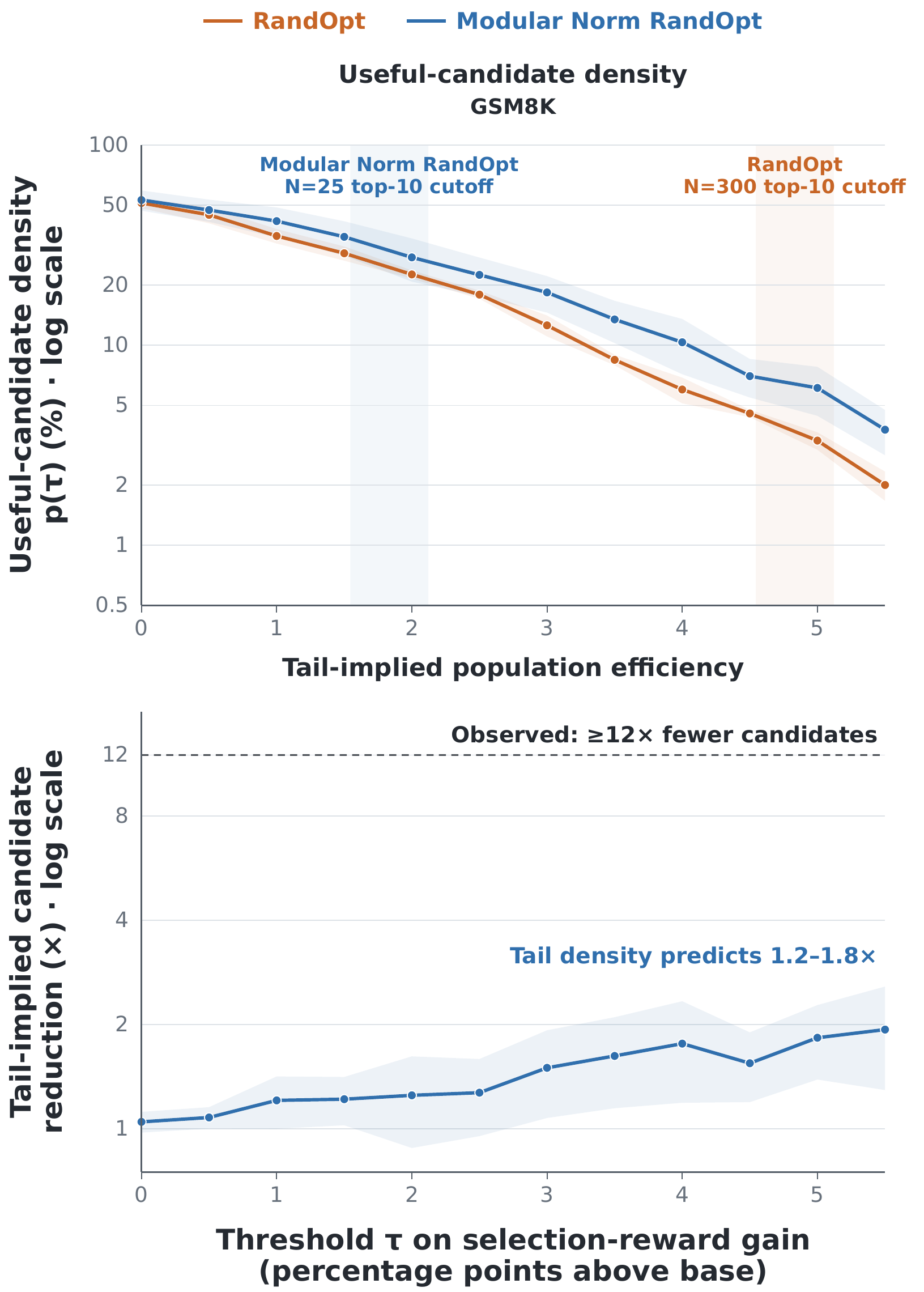}

\end{minipage}
\par

\subsection{Support Redistribution Under Plurality Voting}
\label{sec:support_redistribution}

\begin{figure}[!htbp]
\centering

\begin{minipage}[t]{0.48\linewidth}
\vspace{0pt}
\centering

\includegraphics[width=\linewidth]{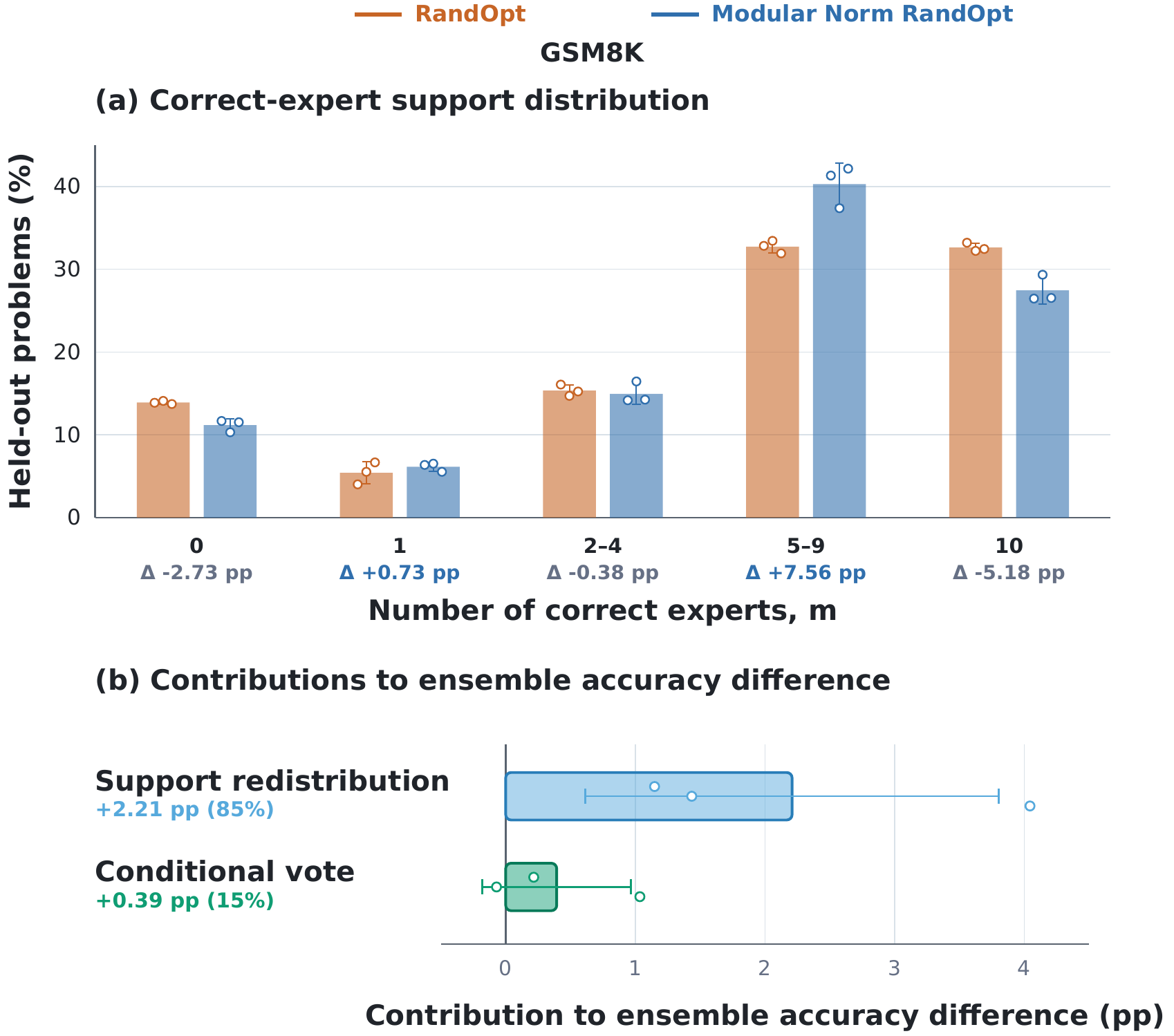}

\end{minipage}%
\hfill%
\begin{minipage}[t]{0.49\linewidth}
\vspace{0pt}
\small
\raggedright

\textbf{Support accounting.}
For method $a$, let $p_a(m)$ be the fraction of held-out evaluation
problems with exactly $m\in\{0,\ldots,K\}$ correct selected experts,
and let $q_a(m)$ be the plurality accuracy conditional on $m$.
The ensemble accuracy is therefore
$\mathrm{Acc}_a=\sum_{m=0}^{K} p_a(m)q_a(m)$.

\par\smallskip
\textbf{Observed shift.}
Modular Norm RandOpt reduces unsupported problems and moves probability
mass toward the bin containing $5$ to $9$ correct experts.
A symmetric decomposition of the $2.60$ point ensemble gain
attributes $2.21$ points, or $85\%$, to support redistribution
and $0.39$ points, or $15\%$, to changes in conditional
plurality accuracy.

\par\smallskip
\textbf{Interpretation.}
The gain is associated primarily with a more favorable distribution
of correct-expert support, rather than improved voting at a fixed
support level.

\end{minipage}

\vspace{-0.2em}

\caption{%
\textbf{Selected-set support on GSM8K.}
Modular Norm RandOpt uses $N=25$, RandOpt uses $N=300$, and both use
$K=10$.
\textbf{(a)} Distribution of held-out evaluation problems by the number
$m$ of selected experts producing the gold answer.
\textbf{(b)} Decomposition of the ensemble-accuracy difference into
support redistribution and conditional voting.
Bars show means, markers denote individual seeds, and error bars show
mean $\pm$ one sample SD over seeds 42--44, not confidence intervals.
}
\label{fig:support_decomposition}

\end{figure}

Tail enrichment does not explain the full population-efficiency gap,
so we next examine the support structure of the selected experts in
Figure~\ref{fig:support_decomposition}.

\FloatBarrier

\section{Limitations and Conclusion}
\label{sec:limitations_conclusion}

\paragraph{Limitations.}
The benefits of Modular Norm RandOpt are task and model dependent
rather than uniform. Cross-family evaluation covers Llama 3.2 $3$B,
Gemma 3 $4$B, and OLMo 3 $7$B-Instruct on Countdown and GSM8K,
with mixed results on OLMo. Evaluation beyond $7$B, additional
architectures, and broader task families remain directions for
future work. We have not fully characterized sensitivity to
calibration choices or radius-selection grids. Our wall-clock
measurements use one model, two tasks, and a single hardware
configuration, so absolute speedups may vary across systems.
Moreover, reducing the search population does not remove the cost
of evaluating the final $K$ selected experts. Finally, the
candidate-tail and support-redistribution analyses are observational,
with the detailed decomposition focused on GSM8K.

\paragraph{Conclusion.}
We introduced \textbf{Modular Norm RandOpt}, which replaces RandOpt's
isotropic candidate distribution with an architecture-aware geometry
based on module-specific natural norms and calibrated modular scales,
while preserving candidate evaluation, top-$K$ selection, and plurality
voting. The method matches or exceeds substantially larger RandOpt
populations using $3\times$ fewer candidates on Countdown and at least
$12\times$ fewer candidates on GSM8K, with corresponding wall-clock
savings. Across seven tasks and three Qwen scales, it outperforms
RandOpt in $14$ of $21$ settings and ties in one, with higher mean
accuracy at every tested scale on Countdown, GSM8K, and MATH-500.
The gains also transfer across model families, improving all four
Llama 3.2 $3$B and Gemma 3 $4$B comparisons, with additional
cross-family evaluation extending to OLMo 3 $7$B. On Qwen2.5-1.5B, our ensembles also achieve higher mean accuracy
than iterative ES, MeZO, and task-adapted ZO-Finetuner on
Countdown and GSM8K at comparable main-run evaluation budgets. Ablations across module coverage, norm choice, and scale construction demonstrate the benefits of the full method. Candidate-tail enrichment alone does not explain the observed
population efficiency; on GSM8K, most of the ensemble gain is
associated with a more favorable distribution of correct-expert
support after selection. Together, these results establish perturbation
geometry as an important design choice for population-efficient,
gradient-free search around pretrained models and show that modular
network structure can be used to construct more useful candidate
ensembles.

\subsection*{AI use statement}
Large language models (LLMs) were used solely for improving written
English and stylistic refinement under careful author supervision.
They were not used to generate scientific content, design experiments,
analyze data, or make intellectual contributions, and did not influence
any reported results.

\subsection*{Ethics statement}

This work uses publicly available language models and benchmark
datasets and does not involve human subjects or private personal
data. We are not aware of any specific ethical concerns beyond
those generally associated with large language models.

\subsection*{Reproducibility statement}

We provide the full method specification and experimental protocol in
Sections~\ref{sec:method}--\ref{sec:experiments}, with additional
implementation details, calibration procedures, hyperparameter settings,
evaluation protocols, and prompt templates in the Appendix.
We also report random seeds and the configurations used for the main
comparisons and ablations.
Code and experiment configurations will be released to facilitate
reproduction of the reported results.

\subsubsection*{Acknowledgments}

We thank Phillip Isola and Yulu Gan for their insightful discussions
and continued feedback, which helped refine our experimental evaluation
and interpretation of the results. We are particularly grateful to
Phillip Isola for suggesting the use of modular norms to scale random
weight perturbations, which helped motivate this work.
We also thank the Ishiguro Laboratory at The University of Osaka
for providing the computational resources used in this research.
This work was partially supported by the JST-SICORP project.

\newpage

\bibliography{iclr2027_conference}
\bibliographystyle{iclr2027_conference}

\clearpage
\appendix

\makeatletter
\@addtoreset{table}{section}
\makeatother
\renewcommand{\thetable}{\thesection\arabic{table}}
\renewcommand{\theHtable}{appendix.\thesection.\arabic{table}}

\setcounter{secnumdepth}{2}

\startcontents[appendix]

\section*{Appendix Contents}

\vspace{0.5em}

\begingroup
\small
\hypersetup{hidelinks}
\printcontents[appendix]{app}{1}[2]{%
    \contentsmargin{24pt}%
}
\endgroup

\clearpage

\section{Sensitivity Calibration and Recursive Modular Scales}
\label{app:sensitivity_calibration}

This appendix specifies the natural norms, architecture-based mass
allocation, sensitivity calibration, and final tensor-wise scales used
to construct the perturbation geometry in
Equation~\ref{eq:modular_perturbation}.

\subsection{Module Norms}
\label{app:module_norms}

Table~\ref{tab:module_norms} specifies the role-specific natural
norm assigned to each parameter type used in our models.
The same role-specific rule is applied in every Transformer layer,
including fused projections and fallback cases.

\begin{table}[H]
\centering
\caption{Leaf-module norms used to normalize sampled parameter perturbations.
For fused QKV and fused gate--up parameters, the norm is evaluated on the
stored physical matrix.}
\label{tab:module_norms}
\small
\renewcommand{\arraystretch}{1.07}
\begin{tabularx}{\linewidth}{@{}lXl@{}}
\toprule
Tensor type & Parameter roles & $\|Z_p\|_{\mathsf{M}_p}$ \\
\midrule
Embedding matrix
& Token embedding
& $\displaystyle \max_j\|Z_{p,j,:}\|_2$ \\
Matrix-valued linear map
& Q, K, V, attention-output, gate, up, down, and separately stored
  output-head weights
& $\displaystyle \sigma_{\max}(Z_p)$ \\
One-dimensional tensor
& Input, post-attention, and final RMSNorm weights; other vector parameters
& $\displaystyle \|Z_p\|_\infty$ \\
Scalar
& Scalar parameters, when present
& $\displaystyle |Z_p|$ \\
Other tensor
& Fallback for unmatched tensors
& $\displaystyle \|Z_p\|_F$ \\
\bottomrule
\end{tabularx}
\end{table}

Matrix spectral norms used in the leaf-module norm are computed by a
deterministic power-iteration routine initialized from an all-ones vector and
therefore introduce no additional random seed.

\subsection{Calibration Protocol and Jacobian Estimation}
\label{app:calibration_protocol}
\label{app:jacobian_estimation}

For each model scale, we use the first $n_{\mathrm{cal}}=64$ prompts, in fixed
order, from the $200$-example Countdown training split constructed with split
seed $42$.  The calibrator performs no additional shuffle or subsampling, and
labels are not used in the Jacobian-norm estimates.

For a model with $L$ Transformer layers, example
$e\in\{0,\ldots,63\}$ is assigned four layers,
\begin{equation}
    \ell_{e,k}
    =
    (4e+k)\bmod L,
    \qquad
    k\in\{0,1,2,3\}.
    \label{eq:app_layer_schedule}
\end{equation}
Let
\begin{equation}
    \mathcal{I}_\ell
    :=
    \{e:\ell_{e,k}=\ell
      \text{ for some }k\in\{0,1,2,3\}\}.
    \label{eq:app_layer_example_set}
\end{equation}
Thus $n_{\mathrm{cal}}=64$ is the number of prompts, not the number of local
measurements per layer.  For Qwen2.5-1.5B ($L=28$), layers $0$--$3$ receive
ten estimates and layers $4$--$27$ receive nine; the final RMSNorm is
evaluated on all $64$ prompts.

For layer $\ell$, let $\mathsf{N}_{\mathrm{in},\ell}$ and
$\mathsf{N}_{\mathrm{post},\ell}$ denote the two RMSNorm maps,
$\mathsf{A}_\ell$ the attention map, and $\mathsf{F}_\ell$ the MLP map.  Define
\begin{align}
    \mathsf{R}_{\mathrm{attn},\ell}(h)
    &:={}
    h+\mathsf{A}_\ell
    \bigl(\mathsf{N}_{\mathrm{in},\ell}(h)\bigr),
    \\
    \mathsf{R}_{\mathrm{mlp},\ell}(r)
    &:={}
    r+\mathsf{F}_\ell
    \bigl(\mathsf{N}_{\mathrm{post},\ell}(r)\bigr),
    \\
    \mathsf{B}_\ell(h)
    &:={}
    \mathsf{R}_{\mathrm{mlp},\ell}
    \bigl(\mathsf{R}_{\mathrm{attn},\ell}(h)\bigr).
    \label{eq:app_functional_maps}
\end{align}
The estimator uses these seven maps in the order
\begin{equation}
    \bigl(
    \mathsf{N}_{\mathrm{in},\ell},
    \mathsf{N}_{\mathrm{post},\ell},
    \mathsf{A}_\ell,
    \mathsf{R}_{\mathrm{attn},\ell},
    \mathsf{F}_\ell,
    \mathsf{R}_{\mathrm{mlp},\ell},
    \mathsf{B}_\ell
    \bigr),
    \label{eq:app_metric_order}
\end{equation}
with corresponding offsets $\kappa(\tau)=0,1,\ldots,6$.

For a functional map $f$ evaluated at activation $x$, define
\begin{equation}
    \gamma(f,x)
    :=
    \max_{\|v\|_2=1}\|J_f(x)v\|_2,
    \qquad
    J_f(x)=\frac{\partial f(x)}{\partial x},
    \label{eq:app_exact_sensitivity}
\end{equation}
where activation tensors are vectorized before applying the Euclidean norm.
For each example, layer, and functional map, we use a dedicated
deterministic random seed to initialize the Gaussian vector for power
iteration. The final RMSNorm uses a separate deterministic seed.

With $g\sim\mathcal{N}(\mathbf{0},\mathbf{I})$ and
$v_0=g/\|g\|_2$, three JVP/VJP power-iteration steps are
\begin{align}
    u_k
    &=
    \frac{J_f(x)v_k}
    {\max\{\|J_f(x)v_k\|_2,10^{-12}\}},
    \\
    v_{k+1}
    &=
    \frac{J_f(x)^\top u_k}
    {\max\{\|J_f(x)^\top u_k\|_2,10^{-12}\}},
    \qquad k=0,1,2,
    \label{eq:app_power_iteration}
\end{align}
followed by
\begin{equation}
    \widehat{\gamma}(f,x)
    =
    \max\{\|J_f(x)v_3\|_2,10^{-12}\}.
    \label{eq:app_power_estimate}
\end{equation}

For each layer and functional map, we aggregate only the examples assigned to
that layer:
\begin{equation}
    \Gamma_{\ell,\tau}
    =
    \exp\!\left[
        Q_{0.9}^{\mathrm{linear}}
        \left(
            \left\{
                \log\max
                \bigl(
                    \widehat{\gamma}_{e,\ell,\tau},
                    10^{-12}
                \bigr)
            \right\}_{e\in\mathcal{I}_\ell}
        \right)
    \right].
    \label{eq:app_aggregated_sensitivity}
\end{equation}
We use the 90th percentile to emphasize upper-tail local
sensitivities while reducing dependence on isolated extreme values.
The final-RMSNorm aggregate applies the
same rule to all $64$ prompts.

\subsection{Mass Allocation and Recursive Base Scales}
\label{app:mass_allocation}

Our mass allocation is motivated by the modular-norm framework,
where user-specified mass fractions bound individual modules'
contributions to linearized output changes relative to the full
modular-norm perturbation, under the well-normedness assumptions
\citep{modular-norm}.
We adopt this allocation principle as a structural prior for
perturbation sampling.
Following the depth-independent mass-taring strategy
\citep{modular-norm}, we assign fixed total masses
to architectural groups and distribute them across repeated layers.
This prevents embedding and output-head mass fractions from
vanishing solely as depth increases.
The values in Table~\ref{tab:group_masses} instantiate this prior
with equal total mass for attention and MLP and smaller total
masses for normalization and other parameters.

\begin{table}[H]
\centering
\caption{Total mass assigned to each parameter group.}
\label{tab:group_masses}
\small
\begin{tabular}{@{}lc@{}}
\toprule
Parameter group $g$ & Total mass $M_g$ \\
\midrule
Embedding & $1$ \\
Attention & $1/2$ \\
MLP & $1/2$ \\
Output head & $1$ \\
Normalization & $1/10$ \\
Other & $1/10$ \\
\bottomrule
\end{tabular}
\end{table}

Let tensor $p$ belong to logical module $u$, parameter group $g$, and repeated
layer instance $\ell$.  Its mass is
\begin{equation}
    m_p
    =
    \frac{M_g}{L_g}
    \frac{w_u}{\sum_{v\in\mathcal{U}_{\ell,g}}w_v}
    \frac{1}{n_u},
    \label{eq:app_parameter_mass}
\end{equation}
where $L_g$ is the number of repeated instances in group $g$,
$\mathcal{U}_{\ell,g}$ is the set of logical modules in that group and
instance, $w_u$ is the logical multiplicity, and $n_u$ is the number of
physical tensors belonging to $u$.  We use $w_u=3$ for fused QKV, $w_u=2$
for fused gate--up, and $w_u=1$ otherwise.

For a sequential composition
$\mathsf{N}=\mathsf{F}_n\circ\cdots\circ\mathsf{F}_1$, the modular rule
propagates the parent scale to child $i$ as
\begin{equation}
    s_{\mathsf{F}_i}
    =
    s_{\mathsf{N}}
    \frac{m_{\mathsf{N}}}{m_{\mathsf{F}_i}}
    \prod_{j>i}\gamma(\mathsf{F}_j).
    \label{eq:app_recursive_scale}
\end{equation}
The downstream-sensitivity product is absent for parallel and residual
composition.  Setting the root scale and all functional sensitivities to one
makes the mass ratios telescope, yielding
\begin{equation}
    s_p^{\mathrm{base}}
    =
    \frac{m_{\mathsf{M}}}{m_p},
    \qquad
    m_{\mathsf{M}}
    =
    \sum_{p\in\mathcal{P}}m_p.
    \label{eq:app_base_scale}
\end{equation}
Tensors with $m_p=0$ are excluded from perturbation.

\subsection{Role-Specific Corrections and Final Scales}
\label{app:raw_correction}

For a linear map
$W\in\mathbb{R}^{d_{\mathrm{out}}\times d_{\mathrm{in}}}$, define
\begin{equation}
    \lambda(W)
    =
    \sqrt{\frac{d_{\mathrm{in}}}{d_{\mathrm{out}}}}
    \sigma_{\max}(W).
    \label{eq:app_linear_gain}
\end{equation}
For fused QKV and fused gate--up projections, the corresponding logical
matrices are concatenated along the output-row dimension before evaluating
Equation~\ref{eq:app_linear_gain}.  We use
\begin{equation}
    \operatorname{clip}(x,a,b)
    :=
    \min\{\max\{x,a\},b\}.
    \label{eq:app_clip}
\end{equation}
For attention in layer $\ell$, define
\begin{align}
    q_\ell
    &:={}
    \operatorname{clip}
    \bigl(\lambda(W_{\mathrm{QKV},\ell}),\frac{1}{4},4\bigr),
    \\
    o_\ell
    &:={}
    \operatorname{clip}
    \bigl(\lambda(W_{O,\ell}),\frac{1}{4},4\bigr),
    \\
    \widetilde{\Gamma}^{A}_\ell
    &:={}
    \operatorname{clip}
    \bigl(\Gamma_{\ell,\mathrm{attention}},\frac{1}{4},4\bigr),
    \\
    \phi_{A,\ell}
    &:={}
    \operatorname{clip}
    \left(
        \frac{\widetilde{\Gamma}^{A}_\ell}{q_\ell o_\ell},
        \frac{1}{4},
        4
    \right).
    \label{eq:app_attention_factors}
\end{align}
For the MLP, define
\begin{align}
    g_\ell
    &:={}
    \operatorname{clip}
    \bigl(\lambda(W_{\mathrm{gate/up},\ell}),\frac{1}{4},4\bigr),
    \\
    d_\ell
    &:={}
    \operatorname{clip}
    \bigl(\lambda(W_{\mathrm{down},\ell}),\frac{1}{4},4\bigr),
    \\
    \widetilde{\Gamma}^{F}_\ell
    &:={}
    \operatorname{clip}
    \bigl(\Gamma_{\ell,\mathrm{MLP}},\frac{1}{4},4\bigr),
    \\
    \phi_{F,\ell}
    &:={}
    \operatorname{clip}
    \left(
        \frac{\widetilde{\Gamma}^{F}_\ell}{g_\ell d_\ell},
        \frac{1}{4},
        4
    \right).
    \label{eq:app_mlp_factors}
\end{align}

\begin{table}[H]
\centering
\caption{Raw correction assigned to each corrected Qwen parameter role.}
\label{tab:raw_corrections}
\small
\begin{tabularx}{0.75\linewidth}{@{}Xl@{}}
\toprule
Physical parameter role in layer $\ell$ & $r_p^{\mathrm{raw}}$ \\
\midrule
Input RMSNorm & $q_\ell\phi_{A,\ell}o_\ell$ \\
Q, K, or V projection, including fused QKV & $\phi_{A,\ell}o_\ell$ \\
Attention output projection & $1$ \\
Post-attention RMSNorm & $g_\ell\phi_{F,\ell}d_\ell$ \\
Gate or up projection, including fused gate--up & $\phi_{F,\ell}d_\ell$ \\
MLP down projection & $1$ \\
\bottomrule
\end{tabularx}
\end{table}

Roles not listed in Table~\ref{tab:raw_corrections} use
$r_p^{\mathrm{raw}}=1$.  Let
$\mathcal{P}_\ell:=\{q\in\mathcal{P}:\ell(q)=\ell\}$.  For each tensor in a
repeated Transformer layer, the final correction and scale are
\begin{equation}
    \rho_p
    =
    \operatorname{clip}
    \left(
        \frac{r_p^{\mathrm{raw}}}
        {\operatorname{median}
         \{r_q^{\mathrm{raw}}:q\in\mathcal{P}_{\ell(p)}\}},
        \frac{1}{2},
        2
    \right),
    \qquad
    s_p=s_p^{\mathrm{base}}\rho_p.
    \label{eq:app_final_scale}
\end{equation}
The $[1/2,2]$ bound prevents sensitivity corrections from
overwhelming the architecture-based baseline.
For active tensors outside the repeated Transformer layers, we set
$\rho_p=1$. The profile is constructed once before candidate search and
remains fixed throughout the population run.

\section{Implementation and Evaluation Details}
\label{app:implementation}

\subsection{Data, Generation, and Scoring Protocol}
\label{app:data_protocol}

\begin{table}[H]
\centering
\small
\setlength{\tabcolsep}{3pt}
\caption{\textbf{Task-specific data, generation, and scoring protocol.}
Selection uses 200 fixed examples for every task.
Completion caps are identical across RandOpt and Modular Norm RandOpt
within each task.}
\label{tab:task_protocol}

\begin{tabular}{lrrrll}
\toprule
Task & Selection & Held-out & Cap &
Selection reward & Held-out metric \\
\midrule
Countdown     & 200 & 500  & 1024 & legal answer + format bonus
              & numeric-answer accuracy \\
GSM8K         & 200 & 1319 & 1024 & normalized-answer accuracy
              & normalized-answer accuracy \\
MBPP          & 200 & 500  & 2048 & all stored tests pass
              & all stored tests pass \\
ROCStories    & 200 & 9817 & 64   & $0.6$ position + $0.4$ adjacency
              & exact five-sentence order \\
USPTO-50K     & 200 & 1001 & 64   & micro accuracy
              & balanced accuracy \\
MATH-500      & 200 & 300  & 2048 & normalized-answer accuracy
              & normalized-answer accuracy \\
OlympiadBench & 200 & 474  & 2048 & normalized-answer accuracy
              & normalized-answer accuracy \\
\bottomrule
\end{tabular}
\end{table}

\paragraph{Task-specific scoring.}
Countdown requires a legal arithmetic expression using every supplied
number exactly once and evaluating to the target.
GSM8K uses normalized final-answer matching.
For MBPP, the prompt displays at most three stored tests, while scoring
executes the complete stored test list and setup code.
ROCStories selection uses a continuous $0.6$ position plus $0.4$
adjacent-order reward, whereas held-out evaluation requires the complete
five-sentence ordering to be exact.
USPTO-50K removes atom-map indices before prompting; candidate selection
uses micro accuracy and held-out reporting uses balanced accuracy.
MATH-500 and OlympiadBench use normalized final-answer matching.

\subsection{Candidate Scoring and Decoding}
\label{app:scoring_decoding}

\paragraph{Candidate scoring.}
For candidate $i$ and selection example $j$, let $r_{ij}$ denote the
task-specific selection reward. Candidates are ranked by
\begin{equation}
    R_i=\frac{1}{M}\sum_{j=1}^{M} r_{ij},
    \qquad M=200.
    \label{eq:candidate_score}
\end{equation}
The top $K$ candidates under $R_i$ are retained for ensembling.
Task-specific definitions of $r_{ij}$ are given in
Table~\ref{tab:task_protocol}.

\paragraph{Decoding and tie breaking.}
All canonical runs use greedy decoding with one completion per prompt.
Invalid or unextractable answers receive no vote.
Candidate-score ties preserve candidate order, and plurality-vote ties
are resolved by the answer first produced by the highest-ranked selected
candidate. RandOpt and Modular Norm RandOpt use identical decoding,
extraction, ranking, and voting procedures.

\paragraph{Generation configuration.}
Canonical Qwen runs use one vLLM engine \citep{kwon2023vllm}
with tensor parallelism one.
The completion caps in Table~\ref{tab:task_protocol} are configured upper
bounds rather than observed generation lengths.
Selection and held-out generation use no prompt truncation, whereas
sensitivity calibration uses only the leading $128$ prompt tokens.

\subsection{Perturbation Settings and Radius Selection}
\label{app:radius_selection}
\label{app:hyperparameter_selection}


\begin{table}[H]
\centering
\captionsetup{
    font=footnotesize,
    skip=3pt
}
\caption{\textbf{Perturbation-scale selection for the primary experiments.}
\textbf{(a)} Adopted values and tested grids.
\textbf{(b)} Qwen scale-selection procedure on Qwen2.5-1.5B Countdown.
Llama, Gemma, and OLMo use the functional-matching protocol in
Table~\ref{tab:functional_matching_protocol}.}
\label{tab:hparam_selection}

\footnotesize
\setlength{\tabcolsep}{4pt}
\renewcommand{\arraystretch}{1.10}


\begin{minipage}{\linewidth}
\centering
{\small\textbf{(a) Adopted scales and tested grids}}
\par\vspace{0.25em}

\begin{tabularx}{\linewidth}{
@{}
>{\raggedright\arraybackslash}p{0.27\linewidth}
>{\centering\arraybackslash}p{0.18\linewidth}
>{\raggedright\arraybackslash}X
@{}
}
\toprule
Setting & Adopted value & Tested grid \\
\midrule

Qwen RandOpt
& $\sigma=0.0005$
& $\{0.0001,\allowbreak
     0.0002,\allowbreak
     0.0005,\allowbreak
     0.001,\allowbreak
     0.002\}$ \\

Qwen MN RandOpt
& $R=0.16$
& $\{0.04,\allowbreak
     0.08,\allowbreak
     0.16,\allowbreak
     0.32,\allowbreak 
     0.64\}$ \\

Llama MN RandOpt
& $R=0.32$
& $\{0.01,\allowbreak
     0.02,\allowbreak
     0.04,\allowbreak
     0.08,\allowbreak
     0.16,\allowbreak
     0.32,\allowbreak
     0.64\}$ \\

Gemma MN RandOpt
& $R=0.64$
& Same as Llama \\

OLMo MN RandOpt
& $R=0.64$
& Same as Llama \\

\bottomrule
\end{tabularx}
\end{minipage}

\par\vspace{0.65em}


\begin{minipage}{\linewidth}
\centering
{\small\textbf{(b) Qwen scale-selection procedure}}
\par\vspace{0.25em}

\begin{tabularx}{\linewidth}{
@{}
>{\raggedright\arraybackslash}p{0.27\linewidth}
>{\raggedright\arraybackslash}X
@{}
}
\toprule
Setting & Protocol \\
\midrule

Trial configuration
& $N=100$, $K=25$ for both methods \\

Candidate ranking
& Same 200 fixed Countdown training examples \\

Scale-selection data
& Same 500-example Countdown development set \\

Tuning seeds
& 39--41 for every scale of both methods \\

Selection criterion
& Highest mean development-set ensemble accuracy over seeds 39--41 \\

Exact tie
& Smaller scale \\

\bottomrule
\end{tabularx}
\end{minipage}

\end{table}

\paragraph{Scale selection.}
For Qwen2.5-1.5B Countdown, we select the RandOpt noise scale
$\sigma$ and the Modular Norm RandOpt radius $R$ using the same
grid-search protocol. For each scale, we use $N=100$, $K=25$,
and seeds 39--41, rank candidates on the same 200 training
examples, and evaluate the selected ensemble on the same
500-example development set. We select the scale with the
highest mean development ensemble accuracy, breaking exact ties
in favor of the smaller scale. This procedure selects
$\sigma=0.0005$ for RandOpt and $R=0.16$ for Modular Norm RandOpt.
The development set is distinct from the primary evaluation set,
and target-task evaluation sets are not used for scale selection.

\paragraph{Functional-matching objective.}
For Llama, Gemma, and OLMo, the radius is chosen to match functional
changes to isotropic RandOpt rather than optimize answer accuracy.
Let $D_{\mathrm{KL}}(R)$ and $D_h(R)$ denote the symmetric next-token
KL divergence and relative hidden-state RMS displacement from the
unperturbed model for Modular Norm RandOpt at radius $R$.
The superscript $\mathrm{iso}$ denotes the corresponding reference
measurements. We minimize
\begin{equation}
    J(R)
    =
    \left[
        \log
        \frac{D_{\mathrm{KL}}(R)}
             {D_{\mathrm{KL}}^{\mathrm{iso}}}
    \right]^2
    +
    \left[
        \log
        \frac{D_h(R)}
             {D_h^{\mathrm{iso}}}
    \right]^2.
    \label{eq:radius_matching_objective}
\end{equation}
Each term is zero at exact matching and penalizes both larger
and smaller relative changes.


\begin{table}[H]
\centering
\captionsetup{
    font=footnotesize,
    skip=3pt
}
\caption{\textbf{Functional-matching protocol.}
Shared by Llama 3.2 $3$B, Gemma 3 $4$B, and OLMo 3 $7$B-Instruct.
The radius grids and selected values are in
Table~\ref{tab:hparam_selection}(a).}
\label{tab:functional_matching_protocol}

\footnotesize
\setlength{\tabcolsep}{4pt}
\renewcommand{\arraystretch}{1.10}

\begin{tabularx}{\linewidth}{
@{}
>{\raggedright\arraybackslash}p{0.27\linewidth}
>{\raggedright\arraybackslash}X
@{}
}
\toprule
Setting & Protocol \\
\midrule

Isotropic reference
& RandOpt with $\sigma=0.0005$ \\

Matching data
& 64 fixed Countdown prompts \\

Perturbation seeds
& First 25 candidate seeds from a 100-candidate pool generated
  with global seed 42 \\

Shared inputs
& Identical prompts and perturbation seeds for every radius
  and the isotropic reference \\

Prompt handling
& Full prompts; fail-fast length limit of 512 tokens \\

Answer generation / labels
& No answer generation or accuracy labels \\

Tie breaking
& Smaller radius for exact ties in $J(R)$ \\

Transfer to GSM8K
& Selected radius and Countdown-calibrated sensitivity profile
  reused unchanged \\

\bottomrule
\end{tabularx}
\end{table}

\subsubsection{Task and Model-Scale Transfer}
\label{app:task_transfer_protocol}

The selected $\sigma$ and $R$ are reused across Qwen model sizes
and target tasks.
The sensitivity profile is calibrated separately for each model size
on Countdown and reused unchanged across target tasks.

\subsection{USPTO-50K Metric}
\label{app:uspto_metric}

In our experiments, USPTO-50K is formulated as ten-class reaction
classification. Candidate selection uses micro accuracy: the fraction
of selection examples assigned the correct class.

Held-out reporting instead uses balanced accuracy, which gives each
class equal weight despite unequal class frequencies:
\begin{equation}
\operatorname{BalancedAcc}
=
\frac{1}{10}\sum_{c=1}^{10}
\frac{
\sum_j \mathbf{1}[y_j=c \land \widehat{y}_j=c]
}{
\sum_j \mathbf{1}[y_j=c]
}.
\label{eq:uspto_balanced_accuracy}
\end{equation}
Here $y_j$ is the reference class and $\widehat{y}_j$ is the ensemble
prediction. All ten classes occur in the evaluation set.
Invalid or missing predictions do not count as correct.

Balanced accuracy is recomputed from the saved ensemble predictions;
it is distinct from the micro accuracy used to rank candidates.

\subsection{Hardware and Runtime Details}
\label{app:hardware}

\begin{table}[H]
\centering
\caption{\textbf{Hardware and evaluation settings for the primary
wall-clock experiments.}
Both methods use 200 selection examples and $K=25$ under the
same hardware and inference settings.}
\label{tab:wallclock_setup}

\footnotesize
\setlength{\tabcolsep}{4pt}
\renewcommand{\arraystretch}{1.10}


\begin{minipage}{\linewidth}
\centering
{\small\textbf{(a) Shared hardware and inference settings}}
\par\vspace{0.25em}

\begin{tabularx}{\linewidth}{
@{}
>{\raggedright\arraybackslash}p{0.28\linewidth}
>{\raggedright\arraybackslash}X
@{}
}
\toprule
Setting & Value \\
\midrule

Model
& Qwen2.5-1.5B-Instruct \\

GPU
& NVIDIA RTX PRO 6000 Blackwell Max-Q (1 GPU per run) \\

vLLM engines per run
& 1 \\

Tensor parallelism
& 1 \\

Weight precision
& \texttt{bfloat16} \\

Decoding
& Greedy \\

Completion cap
& 1,024 tokens \\

\bottomrule
\end{tabularx}
\end{minipage}

\par\vspace{0.65em}


\begin{minipage}{\linewidth}
\centering
{\small\textbf{(b) Population and evaluation sizes}}
\par\vspace{0.25em}

\begin{tabular*}{\linewidth}{@{\extracolsep{\fill}}lrrr@{}}
\toprule
Task & RandOpt $N$ & MN RandOpt $N$ & Evaluation examples \\
\midrule
Countdown & 300 & 100 & 500 \\
GSM8K     & 300 & 25  & 1,319 \\
\bottomrule
\end{tabular*}
\end{minipage}

\end{table}

\paragraph{Timing scope.}
Candidate-search time includes perturbation construction, generation
on the selection set, reward computation, perturbation restoration,
and candidate ranking. For Modular Norm RandOpt, construction of the
fixed scale map is included in search time.

Pipeline end-to-end time additionally includes model, tokenizer, and
data setup, together with prediction and plurality voting by the
selected $K=25$ experts on the evaluation set. CUDA is synchronized
at phase boundaries.

\paragraph{Excluded costs.}
The reused sensitivity-profile calibration, which takes approximately
19 seconds, is excluded from the reported pipeline times.
Prior hyperparameter exploration and an auxiliary unperturbed-model
evaluation are also outside the timed pipeline. 

\paragraph{Speedup aggregation.}
For each task and population seed, speedup is computed as the RandOpt
elapsed time divided by the Modular Norm RandOpt elapsed time for the
corresponding phase. We report the mean and sample standard deviation
of these three ratios over seeds 42--44.

\section{Additional Experimental Results}
\label{app:additional_results}

\subsection{Population and Ensemble-Size Scaling}
\label{app:population_scaling}

We extend the main population comparison to
$K\in\{1,5,10,25\}$.
All runs use Qwen2.5-1.5B-Instruct and nested prefixes
$N\in\{25,50,100,200,300\}$ of a 300-candidate population.

Table~\ref{tab:extended_k_comparison} reports the same cross-budget
contrast at each ensemble size. The candidate-efficiency advantage is
not uniform across $K$: on GSM8K the contrast is negative at $K=1$
but positive from $K=5$ onward.

\begin{table}[H]
\centering
\small
\caption{\textbf{Cross-budget accuracy differences across ensemble sizes.}
Values are mean accuracy differences in percentage points,
Modular Norm RandOpt minus RandOpt, over seeds 42--44.
Countdown compares Modular Norm RandOpt $N=100$ with RandOpt $N=300$;
GSM8K compares Modular Norm RandOpt $N=25$ with RandOpt $N=300$.}
\label{tab:extended_k_comparison}
\begin{tabular}{@{}rrr@{}}
\toprule
$K$ & Countdown & GSM8K \\
\midrule
1  & $+0.00$ & $-1.49$ \\
5  & $+0.07$ & $+2.35$ \\
10 & $+1.53$ & $+2.60$ \\
25 & $+0.93$ & $+3.11$ \\
\bottomrule
\end{tabular}
\end{table}

\subsection{Ablations and Controls}
\label{app:ablations}
\label{app:ablation_hparams}

\begin{table}[H]
\centering
\captionsetup{
    font=footnotesize,
    skip=3pt
}
\caption{\textbf{Settings for the Countdown ablations.}
\textbf{(a)} Shared experimental settings.
\textbf{(b)} Perturbation configurations.
Radii are defined in each configuration's normalization geometry
and are not numerically comparable across different norm choices.}
\label{tab:ablation_hparams}

\footnotesize
\setlength{\tabcolsep}{4pt}
\renewcommand{\arraystretch}{1.10}


\begin{minipage}{\linewidth}
\centering
{\small\textbf{(a) Shared experimental settings}}
\par\vspace{0.25em}

\begin{tabularx}{\linewidth}{
@{}
>{\raggedright\arraybackslash}p{0.32\linewidth}
>{\raggedright\arraybackslash}X
@{}
}
\toprule
Setting & Value \\
\midrule

Model / task
& Qwen2.5-1.5B / Countdown \\

Selection / evaluation examples
& 200 / 500 \\

Population / ensemble size
& $N=100$, $K=25$ \\

Seeds
& 42--44 \\

Decoding
& Greedy \\

Completion cap
& 1,024 tokens \\

\bottomrule
\end{tabularx}
\end{minipage}

\par\vspace{0.65em}


\begin{minipage}{\linewidth}
\centering
{\small\textbf{(b) Perturbation configurations}}
\par\vspace{0.25em}

\begin{tabularx}{\linewidth}{
@{}
>{\raggedright\arraybackslash}p{0.25\linewidth}
>{\centering\arraybackslash}p{0.09\linewidth}
>{\raggedright\arraybackslash}X
@{}
}
\toprule
Configuration & Radius & Perturbed parameters and scaling rule \\
\midrule

MN RandOpt (full)
& $0.16$
& Full active set; module-specific natural norms with calibrated
  modular scales. \\

\addlinespace[0.3em]
Attention only
& $0.16$
& Attention parameters only; retain full-method calibrated scales
  and mask all non-attention perturbations to zero. \\

\addlinespace[0.3em]
MLP only
& $0.16$
& MLP parameters only; retain full-method calibrated scales
  and mask all non-MLP perturbations to zero. \\

\addlinespace[0.3em]
RMSNorm-only scale correction
& $0.16$
& Full active set; recursive base scales with doubled normalization
  denominators for input and post-attention RMSNorm parameters in
  each block, halving their perturbation magnitudes relative to
  the base-scale perturbations.
  All other parameters, including final RMSNorm, retain their
  base scales. \\

\addlinespace[0.3em]
Frobenius baseline
& $0.5$
& Full active set; per-parameter Frobenius-normalized perturbations. \\

\bottomrule
\end{tabularx}

\end{minipage}
\end{table}

For the attention-only and MLP-only controls, scales are not
recomputed over the active subset, and masked perturbations are
not renormalized. Here, $R$ denotes the full-method radius
before masking.

\section{Comparison with Iterative Baselines}
\label{app:iterative_es}

\subsection{Baseline Configurations and Comparison Protocol}
\label{app:es_protocol}
\label{app:zo_baselines}

We adapt iterative ES \citep{ES-Scale}, MeZO \citep{MeZO},
and ZO-Finetuner \citep{ZO-Finetuner} to Countdown and GSM8K.
All use Qwen2.5-1.5B-Instruct, the same 200 training prompts
per task, greedy decoding with a 1,024-token cap, and final
seeds 42--44. Final tests contain 1,500 Countdown and 1,319
GSM8K examples, with scoring as in Appendix~\ref{app:implementation}.

\begin{table}[H]
\centering

\caption{\textbf{Settings and hyperparameter selection for iterative baselines.}
Each method returns one model.
Hyperparameters are selected on Countdown and reused on GSM8K.
ES uses Z-score reward shaping; MeZO and ZO-Finetuner use zero
weight decay.}
\label{tab:es_hparams}
\label{tab:zo_baseline_hparams}

\footnotesize
\setlength{\tabcolsep}{3pt}
\renewcommand{\arraystretch}{1.10}


\begin{minipage}{\linewidth}
\centering
{\small\textbf{(a) Main-run configuration}}
\par\vspace{0.25em}

\begin{tabularx}{\linewidth}{
@{}
>{\raggedright\arraybackslash}p{0.27\linewidth}
*{3}{>{\raggedright\arraybackslash}X}
@{}
}
\toprule
Setting & Iterative ES & MeZO & ZO-Finetuner \\
\midrule

Precision
& BF16
& FP16
& FP16 \\

Main-run updates
& 100
& 1,500
& 1,500 \\

Perturbed models per update
& 30
& 2
& 2 \\

Perturbation scale
& $\sigma=0.001$
& $\epsilon=0.001$
& $\epsilon=0.001$ \\

Update coefficient / learning rate
& $\alpha=0.0005$
& $10^{-6}$
& $10^{-6}$ \\

Reported checkpoint
& Iteration 100
& Best development
& Best development \\

Generator preparation
& None
& None
& Once, shared \\

\bottomrule
\end{tabularx}
\end{minipage}

\par\vspace{0.65em}


\begin{minipage}{\linewidth}
\centering
{\small\textbf{(b) Hyperparameter selection on Countdown}}
\par\vspace{0.25em}

\begin{tabularx}{\linewidth}{
@{}
>{\raggedright\arraybackslash}p{0.27\linewidth}
*{3}{>{\raggedright\arraybackslash}X}
@{}
}
\toprule
Setting & Iterative ES & MeZO & ZO-Finetuner \\
\midrule

Tuned parameter
& Noise scale $\sigma$
& Learning rate
& Learning rate \\

Search grid
& $\{0.0005,\allowbreak 0.001,\allowbreak 0.002\}$
& $\{10^{-6},\allowbreak 10^{-5},\allowbreak 10^{-4}\}$
& $\{10^{-8},\allowbreak 10^{-7},\allowbreak 10^{-6}\}$ \\

Updates per trial
& 10
& 500
& 500 \\

Tuning seeds
& 39--41
& 39
& 39 \\

Selection criterion
& Highest mean development accuracy
& Best stable trial by development accuracy
& Best stable trial by development accuracy \\

\bottomrule
\end{tabularx}
\end{minipage}

\end{table}

\paragraph{Development sets and checkpoint selection.}
Countdown selection uses the historical 500-question development set.
For GSM8K, MeZO and ZO-Finetuner use 500 development examples
held out from the training pool, disjoint from the fixed training
prompts and final test.
Both methods evaluate development performance at step 0 and every
100 updates, then evaluate the best checkpoint once on the final test.
FP16 is chosen using training-only numerical-stability checks.
No final-test scores are used for selection.

\paragraph{ZO-Finetuner preparation.}
The perturbation generator is trained once on 196 solved
Countdown training examples for 15 epochs, using batch size 4,
learning rate 0.01, and FP32.
This preparation uses supervised solutions and gradients.
Generator weights are then frozen and reused across both tasks
and all seeds, while predicted perturbation amplitudes remain
state-dependent. MeZO requires no such preparation.
Auxiliary costs are reported separately in
Appendix~\ref{app:es-budget}.

\subsection{Evaluation-Budget Accounting}
\label{app:es-budget}

Let $S$ denote the number of final-test examples.
One model--prompt evaluation generates and scores one response
from one model on one prompt.
The main-run budget includes search or training, checkpoint
selection, and final evaluation.
Preparation and hyperparameter-selection costs are reported
separately.
For Modular Norm RandOpt with population size $N$
and ensemble size $K$, the main-run evaluation count is
\begin{equation}
    B_{\mathrm{MN}}(N,K,S)=200N+KS.
    \label{eq:mn_total_eval_budget}
\end{equation}
Iterative ES evaluates $3{,}000$ candidates on the same
$200$ selection prompts and returns one final model, giving
\begin{equation}
    B_{\mathrm{ES}}(S)=200\times 3000+S.
    \label{eq:es_total_eval_budget}
\end{equation}
For MeZO and ZO-Finetuner, $1{,}500$ two-sided updates on $200$
training examples require $600{,}000$ evaluations.
Development evaluation at steps $0,100,\ldots,1500$
uses $16\times500=8{,}000$ additional evaluations.
The selected checkpoint is evaluated once on the final test:
\begin{equation}
    B_{\mathrm{MeZO}}(S)=B_{\mathrm{ZO}}(S)
    =2\times1500\times200+16\times500+S
    =608{,}000+S.
    \label{eq:zo_total_eval_budget}
\end{equation}

We report
\begin{equation}
    \mathrm{Eval.\ red.}
    =\frac{B_{\mathrm{ES}}(S)}{B_{\mathrm{method}}(S)}.
    \label{eq:iterative_eval_reduction}
\end{equation}

This gives $601{,}500/609{,}500$ on Countdown and
$601{,}319/609{,}319$ on GSM8K, both rounded to $0.99\times$.
The ES budget excludes intermediate test monitoring that was
not used to select the reported iteration-100 checkpoint.
These additional calls total $6S$ per run.
In contrast, MeZO and ZO-Finetuner development evaluations are included
because they determine the reported checkpoint.

For Countdown, $S=1500$ and $N=2820$ gives
$B_{\mathrm{MN}}=B_{\mathrm{ES}}=601{,}500$.
For GSM8K, $S=1319$ and $N=2841$ gives
$B_{\mathrm{MN}}=601{,}175$, compared with
$B_{\mathrm{ES}}=601{,}319$.
The $144$-evaluation difference arises from rounding the integer
candidate population downward; both are displayed as $1.00\times$
at the precision used in Table~\ref{tab:iterative-es}.

For Countdown, with $S=1500$, Modular Norm RandOpt with $N=100$
and $K=25$ uses
$200\times100+25\times1500=57{,}500$ main-run model--prompt
evaluations, compared with $609{,}500$ for either MeZO or
ZO-Finetuner. This corresponds to approximately $10.6\times$
fewer evaluations, or a $90.6\%$ reduction.

For GSM8K, Modular Norm RandOpt with $N=100$ and $K=25$ uses
$200\times100+25\times1319=52{,}975$ main-run model--prompt
evaluations, compared with $609{,}319$ for either MeZO or
ZO-Finetuner. This corresponds to approximately $11.5\times$
fewer evaluations, or a $91.3\%$ reduction.

\paragraph{Auxiliary costs.}
Outside the main-run budget, each MeZO or ZO-Finetuner
learning-rate sweep uses 600,000 training and 9,000 development
evaluations; precision checks add 3,200 evaluations per method.
ZO-Finetuner generator preparation costs approximately 623 seconds
once, shared across tasks and seeds.
\subsection{Full Results and Single-Model Controls}
\label{app:es_results}

Both methods perform $3{,}000$ candidate evaluations with
$200$ selection prompts per candidate, corresponding to
$600{,}000$ search model--prompt evaluations.
Iterative ES uses $30$ candidates per iteration for $100$
iterations, whereas Modular Norm RandOpt samples one population
of $3{,}000$ candidates. The $K=1$ and $K=25$ Modular Norm RandOpt results use
the same saved search.

\begin{table}[H]
\centering
\small
\caption{\textbf{Iterative ES and Modular Norm RandOpt at the same
search-evaluation count.}
Accuracy (\%) is mean $\pm$ sample SD over seeds 42--44.
Evaluation uses 1,500 Countdown examples and 1,319 GSM8K examples.
The Countdown evaluation set differs from the 500-example set used
in the primary transfer and population-scaling experiments.}
\label{tab:es_full_endpoints}
\begin{tabular}{@{}lrrr@{}}
\toprule
Method & $K$ & Countdown & GSM8K \\
\midrule
Iterative ES & 1  & $35.67\pm5.08$ & $73.11\pm0.52$ \\
MN RandOpt   & 1  & $16.18\pm1.08$ & $64.42\pm0.29$ \\
MN RandOpt   & 25 & $38.44\pm1.20$ & $74.45\pm0.95$ \\
\bottomrule
\end{tabular}
\end{table}

Modular Norm RandOpt has higher mean accuracy with $K=25$ but
lower accuracy as a single selected model. Matching the search
evaluation count does not equalize inference cost: a $K=25$
ensemble requires $25$ model predictions per input, whereas
iterative ES returns one model. Nor does an equal model--prompt
count imply equal runtime or generated-token count.

\section{Additional Candidate and Ensemble Analyses}
\label{app:additional_analysis}

\subsection{Solution Density on Countdown and GSM8K}
\label{app:solution_density}

This analysis evaluates individual perturbed candidates rather than
selected ensembles. For candidate accuracy $a_i$ and unperturbed-model
accuracy $a_0$, measured as fractions on the same fixed set of
questions, define
\begin{equation}
\delta(t)
=
\frac{1}{N}
\sum_{i=1}^{N}
\mathbf{1}\!\left[100(a_i-a_0)\geq t\right].
\label{eq:solution_density}
\end{equation}
The threshold $t$ is measured in percentage points. For example,
$\delta(2)$ is the fraction of candidates improving over the
unperturbed model by at least two percentage points.

We evaluate 3,000 candidates per method and population seed using
Qwen2.5-1.5B-Instruct, RandOpt with $\sigma=0.0005$, and Modular Norm
RandOpt with $R=0.16$, over seeds 42--44.
The fixed density-evaluation sets contain 500 Countdown questions and
1,000 GSM8K questions. The GSM8K density set is distinct from the
1,319-example set used for the primary ensemble evaluation.

\begin{table}[H]
\centering
\small
\caption{\textbf{Solution densities.}
Values are percentages of candidates averaged over three population
seeds. Thresholds denote accuracy improvements in percentage points,
not relative percentage gains.}
\label{tab:solution_densities}
\begin{tabular}{@{}llrrrr@{}}
\toprule
Task & Method & $t=1$ & $t=2$ & $t=3$ & $t=5$ \\
\midrule
Countdown & RandOpt & 30.88 & 15.98 & 6.71 & 0.43 \\
          & MN RandOpt & 16.92 & 9.61 & 4.76 & 0.81 \\
GSM8K     & RandOpt & 22.20 & 9.56 & 3.06 & 0.13 \\
          & MN RandOpt & 22.99 & 12.28 & 5.36 & 0.41 \\
\bottomrule
\end{tabular}
\end{table}

The ordering depends on task and threshold. On Countdown, Modular Norm
RandOpt has lower density at thresholds of 1--3 points but higher density
at 5 points; on GSM8K, its mean density is higher at all four displayed
thresholds.

These measurements use the adopted perturbation scales and do not match
methods for output-distribution displacement. They neither determine
ensemble accuracy nor imply that Modular Norm RandOpt uniformly produces
more individually useful candidates.

\subsection{Tail-Implied Population Efficiency}
\label{app:tail_efficiency}
Let $p_{a,s}(\tau)$ denote the empirical fraction of candidates from
method $a$ and population seed $s$ whose selection-reward gain over the
pretrained model is at least threshold $\tau$. Each estimate uses the
corresponding $N_{\max}=300$ population.
Under a constant-hit-rate binomial model, define
\begin{equation}
N^{\mathrm{req}}_{a,s}(\tau)
=\min\left\{N\in\mathbb{N}:
\Pr\!\left[\operatorname{Binomial}(N,p_{a,s}(\tau))\geq10\right]
\geq0.9\right\}.
\label{eq:tail_required_population}
\end{equation}
The tail-implied candidate-reduction factor is computed separately for
each paired seed as
\begin{equation}
\rho_s(\tau)
=\frac{N^{\mathrm{req}}_{\mathrm{RandOpt},s}(\tau)}
       {N^{\mathrm{req}}_{\mathrm{MN},s}(\tau)}.
\label{eq:tail_implied_efficiency}
\end{equation}

Figure~\ref{fig:tail_efficiency} plots the mean of these ratios over
seeds $42$, $43$, and $44$, with a band of one sample SD on either side
(denominator $3-1$).
The top-panel densities and selection-cutoff bands use the same
mean-$\pm$-sample-SD convention. Displayed bands are not confidence
intervals and are clipped to the plotting limits where necessary.
Across thresholds $\tau=1.5,2.0,\ldots,5.0$ percentage points, spanning
the observed top-$10$ cutoffs, the mean ratios range from $1.2$ to
$1.8\times$ after rounding.

This diagnostic matches the modeled probability of obtaining at least
$10$ candidates above a single selection-reward threshold. It does not
match the full selected reward distribution, coverage of evaluation
questions, or plurality-vote accuracy. The gap between the
tail-implied $1.2$--$1.8\times$ reduction and the observed population
reduction therefore indicates that tail abundance alone does not
explain the population-efficiency gain.

\subsection{Support Redistribution and Conditional Voting}
\label{app:support_decomposition}

For each held-out evaluation question, let
$m\in\{0,\ldots,K\}$ denote the number of selected experts producing
the correct answer. Let $p_a(m)$ denote the fraction of questions
with support $m$ under method $a$, and let $q_a(m)$ denote plurality
accuracy conditional on that support. Ensemble accuracy is therefore
\begin{equation}
\mathrm{Acc}_a
=
\sum_{m=0}^{K} p_a(m)q_a(m).
\end{equation}

For Modular Norm RandOpt ($M$) and RandOpt ($R$), we use the symmetric
decomposition
\begin{align}
\Delta\mathrm{Acc}
&=
\frac{1}{2}
\sum_m
\bigl[p_M(m)-p_R(m)\bigr]
\bigl[q_M(m)+q_R(m)\bigr]
\\
&\quad+
\frac{1}{2}
\sum_m
\bigl[p_M(m)+p_R(m)\bigr]
\bigl[q_M(m)-q_R(m)\bigr].
\label{eq:support_symmetric_decomposition}
\end{align}
The first term measures the contribution associated with redistribution
of support across questions, while the second measures changes in
plurality accuracy conditional on support.

We analyze GSM8K at $K=10$, comparing Modular Norm RandOpt with
$N=25$ against RandOpt with $N=300$.
Predictions reconstructed from the saved selected-expert outputs match
the recorded ensemble results.

\begin{table}[H]
\centering
\small
\caption{\textbf{Per-seed decomposition of the GSM8K ensemble-accuracy
difference.}
Contributions are Modular Norm RandOpt minus RandOpt in percentage
points. Values are rounded independently, so displayed components may
not sum exactly.}
\label{tab:support_seed_accounting}
\begin{tabular}{@{}lrrr@{}}
\toprule
Seed & Support redistribution & Conditional voting & Total \\
\midrule
42 & 1.15 & 0.22 & 1.36 \\
43 & 1.44 & $-0.07$ & 1.36 \\
44 & 4.04 & 1.04 & 5.08 \\
\midrule
Mean & 2.21 & 0.39 & 2.60 \\
\bottomrule
\end{tabular}
\end{table}

Approximately $85\%$ of the mean accuracy difference is assigned to
the support-redistribution term in this accounting.
Across all three seeds, the support-redistribution contribution is
positive and larger than the conditional-voting contribution,
indicating that the ensemble gain is associated primarily with a more
favorable distribution of correct-expert support.

\section{Prompt Templates}
\label{app:prompts}

The boxes below show the exact message content used in all reported
experiments before application of each model's native chat template.
Braced fields denote per-example substitutions. RandOpt and Modular
Norm RandOpt use identical prompts and rendering procedures within
each task.

\begin{promptbox}{Countdown}{PromptCountdown}{PromptCountdownBG}
System message:

You are a helpful assistant. You first think about the reasoning process in your mind and then provide the user with the answer.

User template:

Using the numbers {numbers}, create an equation that equals {target}. You can use basic arithmetic operations (+, -, *, /) and each number can only be used once. Show your work in <think> </think> tags. And return the final answer in <answer> </answer> tags, for example <answer> (1 + 2) / 3 </answer>.

Note:

{numbers} is the supplied number multiset and {target} is the requested value.
\end{promptbox}

\begin{promptbox}{GSM8K}{PromptGSM}{PromptGSMBG}
System message:

[No explicit system-role message]

User template:

{question} Let's think step by step and output the final answer after "####".
\end{promptbox}

\begin{promptbox}{MBPP}{PromptCode}{PromptCodeBG}
System message:

You are a Python programming assistant. Write clean, correct Python code to solve the given problem.

User template:

{text}

Your code should pass these tests:
{tests}

Think through your solution in <think> </think> tags.
Return your final Python code in <answer> </answer> tags, e.g.:
<answer>
def solution(x):
    return x + 1
</answer>

Note:

The prompt displays at most the first three stored tests; scoring executes the complete stored test list and setup code.
\end{promptbox}

\begin{promptbox}{ROCStories}{PromptStory}{PromptStoryBG}
System message:

You are a helpful assistant that excels at story comprehension and logical reasoning. Given shuffled sentences from a story, you carefully analyze the narrative flow and temporal cues to determine the correct chronological order.

User template:

Below are 5 sentences from a story, but they are in the wrong order.
Please arrange them in the correct chronological order.

Title: {title}

Sentence A: {sentence_0}
Sentence B: {sentence_1}
Sentence C: {sentence_2}
Sentence D: {sentence_3}
Sentence E: {sentence_4}

Output the correct order as comma-separated letters (e.g., B,A,D,E,C).
Only output the letters, nothing else.
\end{promptbox}

\begin{promptbox}{USPTO-50K}{PromptChem}{PromptChemBG}
System message:

You are an expert organic chemist. Your task is to classify chemical reactions into one of 10 standard reaction categories based on the transformation type.

## Reaction Classes:
1: Heteroatom alkylation/arylation - N, O, S attacking C (e.g., SN2, ether formation)
2: Acylation - Forming C=O bonds with N, O, S (e.g., amide, ester formation)
3: C-C bond formation - New C-C bonds (e.g., Suzuki, Heck, Grignard)
4: Heterocycle formation - Creating rings with N, O, S
5: Protections - Adding protecting groups (Boc, Bn, TBS, etc.)
6: Deprotections - Removing protecting groups
7: Reductions - Adding H, removing O (e.g., ketone→alcohol, nitro→amine)
8: Oxidations - Adding O, removing H (e.g., alcohol→ketone)
9: Functional group interconversion - Changing one FG to another
10: Functional group addition - Adding new FG to molecule (e.g., halogenation)

User template:

Classify this reaction:

Reactants >> Product:
{rxn_smiles}

Analyze the key transformation and output the class number (1-10) in <answer>X</answer> tags.

Note:

The reaction string is inserted after atom-map indices have been removed.
\end{promptbox}

\begin{promptbox}{MATH-500}{PromptMath500}{PromptMath500BG}
System message:

[No explicit system-role message]


User template:

{problem} Please reason step by step, and put your final answer within \boxed{}.
\end{promptbox}

\begin{promptbox}{OlympiadBench}{PromptOlympiad}{PromptOlympiadBG}
System message:

[No explicit system-role message; the handler constructs the user message.]

User template:

{question}

Let’s think step by step and output the final answer after ####.
\end{promptbox}

\end{document}